\documentclass[10pt, twocolumn, twoside]{IEEEtran}

\usepackage{times}
\usepackage{epsfig}
\usepackage{graphicx}
\usepackage{amsmath}
\usepackage{amssymb}

\usepackage{array}
\usepackage{color}
\usepackage{diagbox}
\usepackage{multirow}
\usepackage{soul}

\newcommand{\tabincell}[2]{\begin{tabular}{@{}#1@{}}#2\end{tabular}}
\newcommand{\mhy}[1]{\textcolor[rgb]{0.0,0.0,0.0}{#1}}

\newcommand{\add}[1]{\textcolor[rgb]{0.0,0.0,0.0}{#1}}

\newcommand{\red}[1]{\textcolor[rgb]{1,0,0}{#1}}
\newcommand{\newylt}[1]{\textcolor[rgb]{0,0,0}{#1}}
\newcommand{\final}[1]{\textcolor[rgb]{0,0,0}{#1}}
\newcommand{\green}[1]{\textcolor[rgb]{0,1,0}{#1}}
\newcommand{\blue}[1]{\textcolor[rgb]{0,0,1}{#1}}
\newcommand{\ie}{\emph{i.e.}} 
\newcommand{\eg}{\emph{e.g.}}

\ifCLASSOPTIONcompsoc
  \usepackage[nocompress]{cite}
\else
  \usepackage{cite}
\fi

\ifCLASSINFOpdf
\else

\fi

\begin{document}
\bstctlcite{IEEEexample:BSTcontrol}

\title{Progressive Glass Segmentation}

\author{Letian~Yu*, Haiyang~Mei*, Wen~Dong, Ziqi~Wei, Li~Zhu, Yuxin~Wang, and~Xin~Yang\textsuperscript{$\dagger$}%
        
\IEEEcompsocitemizethanks{
\IEEEcompsocthanksitem Letian~Yu, Haiyang~Mei, Wen~Dong, Yuxin~Wang, Xin~Yang are with the School of Computer Science and Technology at Dalian University of Technology, Dalian, 116024, China. E-mail: letianyu@mail.dlut.edu.cn; mhy666@mail.dlut.edu.cn; dongwen@mail.dlut.edu.cn; wyx@dlut.edu.cn; xinyang@dlut.edu.cn.
\IEEEcompsocthanksitem Ziqi~Wei is with the School of Software at Tsinghua University, Beijing, 100084, China. E-mail: weizq\_ruc@foxmail.com.
\IEEEcompsocthanksitem Li~Zhu is with the School of Control Science and Engineering at Dalian University of Technology, Dalian, 116024, China. E-mail: zhuli@dlut.edu.cn. 
\IEEEcompsocthanksitem * Letian~Yu and Haiyang~Mei are the joint first authors.
\IEEEcompsocthanksitem \textsuperscript{$\dagger$} Ziqi~Wei and Xin~Yang are the corresponding authors.
}}% <-this % stops an unwanted space
% \thanks{Manuscript received April 19, 2005; revised August 26, 2015.}}%

\markboth{IEEE TRANSACTIONS ON IMAGE PROCESSING}%
{Yu \MakeLowercase{\textit{et al.}}: Progressive Glass Segmentation}

\IEEEtitleabstractindextext{%
\begin{abstract}Glass is very common in the real world. Influenced by the uncertainty about the glass region and the varying complex scenes behind the glass, the existence of glass poses severe challenges to many computer vision tasks, making glass segmentation as an important computer vision task. Glass does not have its own visual appearances but only transmit/reflect the appearances of its surroundings, making it fundamentally different from other common objects.
To address such a challenging task, existing methods typically explore and combine useful cues from different levels of features in the deep network. As there exists a characteristic gap between level-different features, \textit{i.e.}, deep layer features embed more high-level semantics and are better at locating the target objects while shallow layer features have larger spatial sizes and keep richer and more detailed low-level information, fusing these features naively thus would lead to a sub-optimal solution. 
In this paper, we approach the effective features fusion towards accurate glass segmentation in two steps. First, we attempt to bridge the characteristic gap between different levels of features by developing a Discriminability Enhancement (DE) module which enables level-specific features to be a more discriminative representation, alleviating the features incompatibility for fusion. Second, we design a Focus-and-Exploration Based Fusion (FEBF) module to richly excavate useful information in the fusion process by highlighting the common and exploring the difference between level-different features.
Combining these two steps, we construct a \textbf{P}rogressive \textbf{G}lass \textbf{S}egmentation \textbf{Net}work (\textbf{PGSNet}) which uses multiple DE and FEBF modules to progressively aggregate features from high-level to low-level, implementing a coarse-to-fine glass segmentation.
In addition, we build the first home-scene-oriented glass segmentation dataset for advancing household robot applications and in-depth research on this topic.
Extensive experiments demonstrate that our method outperforms 26 cutting-edge models on three challenging datasets under four standard metrics. 
The code and dataset will be made publicly available.
\end{abstract}

}

\maketitle

\IEEEdisplaynontitleabstractindextext

\IEEEpeerreviewmaketitle

\section{Introduction}

\mhy{\IEEEPARstart{G}{lass} is commonly presented in daily-life scenes, \eg, window panes, glass doors/walls, and glass guardrails, as practical and decorative usages. Yet, glass confuses many vision systems due to its transparency and invisibility \cite{whelan2018reconstructing,Mei_2020_CVPR,xie2020segmenting,xie2021segmenting}. \newylt{For example, a drone may crash into glass surfaces without the ability to sense the presence of glass, and the navigation robots also need to avoid bumping into the glass-like objects. Thus, the ability to segment glass is essential for AI agents in many practical applications. } Automatic glass segmentation is a challenging task as glass does not have its own visual appearances but only transmit/reflect the appearances of its surroundings and thus has no relatively fixed semantics/patterns, salient features, or contrastive features \cite{yang2019mirrornet}, making it fundamentally different from other common/salient/mirror object detection/segmentation problems.}

% \mhy{Just as human beings typically leverage different cues to identify the existence of glass, existing glass/transparency segmentation methods mainly focus on the exploration of useful information from different levels of features in the deep network. For example, Mei \etal \cite{Mei_2020_CVPR} harvest both high-level and low-level contextual features in a large field-of-view. Xie \etal \cite{xie2020segmenting} leverage the boundary cues. And Xie \etal \cite{xie2021segmenting} model contextual dependencies in a global receptive field via self-attention in the transformer \cite{vaswani2017attention}.}

\begin{figure}[tbp]
	\begin{center}
		\includegraphics[width=1\linewidth]{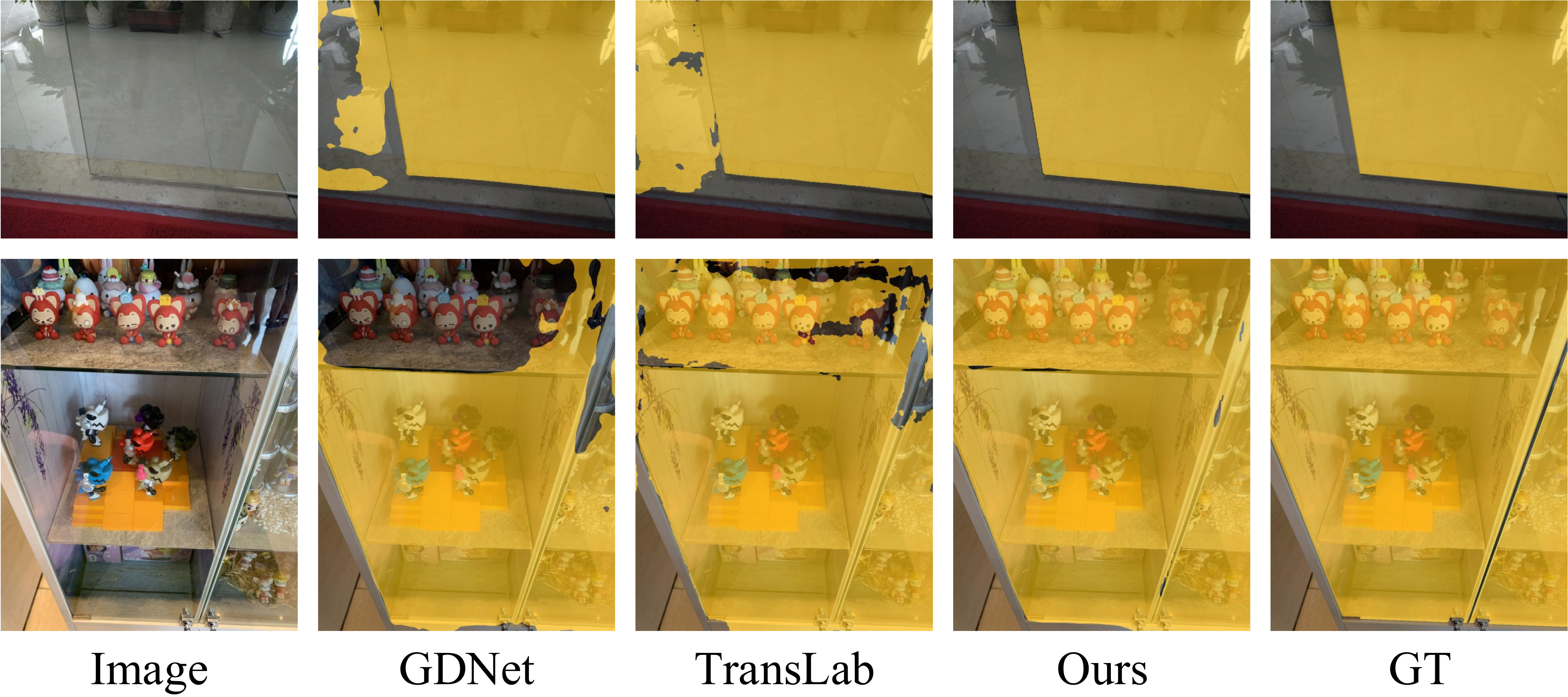}
	\end{center}
	\vspace{-12pt}
	\caption{\mhy{Visual examples of glass segmentation. While the state-of-the-art method GDNet \cite{Mei_2020_CVPR} and TransLab \cite{xie2020segmenting} explore large-field contextual features and boundary cues, respectively, the simple fusion between level-different features (\ie, element-wise multiplication or channel-wise concatenation) ignores the characteristic gap between features, and thus leading to a sub-optimal solution (2\textit{nd} and 3\textit{rd} columns). In contrast, by adopting our proposed features fusion strategy, our method is able to accurately segment the glass (4\textit{th} column).}}
	\label{fig:teaser}
\end{figure}

\mhy{Just as human beings typically leverage different cues to identify the existence of glass, existing glass/transparency segmentation methods mainly focus on the exploration of useful information from different levels of features in the deep network, \eg, both high-level and low-level large-field contextual features \cite{Mei_2020_CVPR}, boundary cues \cite{xie2021segmenting}, and transformer \cite{vaswani2017attention} based global contexts \cite{xie2021segmenting}.}
% For example, Mei \etal \cite{Mei_2020_CVPR} harvest both high-level and low-level contextual features in a large field-of-view. Xie \etal \cite{xie2020segmenting} leverage the boundary cues. And Xie \etal \cite{xie2021segmenting} model contextual dependencies in a global receptive field via self-attention in the transformer \cite{vaswani2017attention}.}
%
\mhy{However, the obtained cues/features typically vary in level and characteristic, \ie, deep layer features in the network usually embed more high-level semantic knowledge and contextual relations and are better at locating the target objects while shallow layer features have larger spatial sizes and keep richer and more detailed low-level information such as edges, lines, and corners, and thus naively combine them (\eg, channel-wise concatenation or element-wise multiplication/addition) for final segmentation would lead to a sub-optimal solution (Figure \ref{fig:teaser}, 2\textit{nd} and 3\textit{rd} columns).}

\mhy{In this paper, we emphasize the \textbf{\textit{significance of the effective fusion}} between different levels of features towards accurate glass segmentation. Specifically, we approach the effective level-different features fusion in two steps. First, we attempt to bridge the characteristic gap between different levels of features by developing a Discriminability Enhancement (DE) module which enables level-specific features to be a more discriminative representation, alleviating the features incompatibility for fusion. Second, we design a Focus-and-Exploration Based Fusion (FEBF) module to achieve the goal of richly excavating useful information by simultaneously highlighting the common and exploring the difference between level-different features in the fusion process. Combining these two steps, we construct a \textbf{P}rogressive \textbf{G}lass \textbf{S}egmentation \textbf{Net}work (\textbf{PGSNet}) which combines multiple DE and FEBF modules to progressively aggregate features from high-level to low-level, implementing a coarse-to-fine glass segmentation. In addition, we build the first home-scene-oriented glass segmentation dataset for advancing household robot applications and in-depth research on this topic. 
	%
	% Extensive experiments demonstrate that our method outperforms 23 cutting-edge detection/segmentation models on three challenging datasets under four standard metrics.
}
\mhy{To sum up, the main contributions of this work are as follows:}
\begin{itemize}
	\item \mhy{First, we develop a novel features fusion strategy which first bridges the characteristic gap between level-different features by the proposed discriminability enhancement (DE) module and then fuses features via the well-designed focus-and-exploration based fusion (FEBF) module.}
	\item \mhy{Second, we construct a progressive glass segmentation network (PGSNet), which leverages our proposed features fusion strategy to progressively aggregate features from high-level to low-level, implementing a coarse-to-fine glass segmentation.}
	\item \mhy{Third, we build the first home-scene-oriented (HSO) glass segmentation dataset, laying a foundation for further research and household robot applications.}
	\item \mhy{Fourth, we achieve state-of-the-art glass segmentation performance on our HSO dataset as well as existing two benchmark datasets. Experimental results demonstrate the effectiveness of our method.}
\end{itemize}

\section{Related Work \\}

\mhy{In this section, we first briefly review state-of-the-art detection/segmentation methods in different fields, including semantic segmentation, salient object detection, as well as specific region segmentation, and then discuss some representative works on deep feature fusion.}

\subsection{Semantic Segmentation}
\mhy{Semantic Segmentation classifies and assigns a semantic label to each pixel in an image. Recently, great progress has been achieved benefited by the advances of deep neural networks. Based on fully convolutional networks (FCNs) \cite{long2015fully}, state-of-the-art methods focus on the exploration of contextual information in different scales/levels and forms \cite{chen2017deeplab,Fu_2019_CVPR_DANet,Huang_2019_ICCV_CCNet,Yang_2018_CVPR_DenseASPP,Yu_2018_ECCV_BiSeNet,Zhang_2018_CVPR_enc,zhao2017pyramid,Zhao_2018_ECCV_psanet, li2020semantic, li2020gated, huang2021fapn}. \newylt{A more comparative survey \cite{minaee2021image} about methods and performances for segmentation has been proposed, grouped into 10 categories, for example, encoder-decoder based models, multi-scale and pyramid network based models and attention-based models. } The glass segmentation problem we strive to address differs from semantic segmentation in that the large intra-class variation, \ie, arbitrary categories of objects could appear behind the glass. Hence, treating glass as an additional semantic category fails to produce satisfactory results as the visible glass content is further semantically classified~\cite{Mei_2020_CVPR}. In this paper, we also exploit different levels of features, but further employ a novel level-different features fusion strategy for accurate glass segmentation.}

% \ylt{The main difference between semantic segmentation and glass segmentation is the category consistency}. \ylt{Different semantic categories can fall into the same glass region, while the same category can be departed by the same glass region.} 
\subsection{Salient Object Detection}
\mhy{Salient Object Detection (SOD) aims to identify the most visually distinctive object(s) in an image of a scene. Hundreds of SOD methods have been proposed in the past decades. Early methods mainly rely on the handcrafted low-level features and heuristic priors such as color \cite{achanta2009frequency} and contrast \cite{cheng2014global}. These features, however, have limited capability to distinguish the salient and non-salient objects, thus the approaches based on them often fail in complex scenes. Recently, state-of-the-art solutions employ convolutional neural networks (CNNs) to exploit different learning strategies and cues such as multi-level feature exploration~\cite{Lee_2016_CVPR,HouPami19Dss,Zhang_2017_ICCV,Zhao_2019_CVPR_pfan,Pang_2020_CVPR_minet,Mei_2021_TCSVT}, recurrent and iterative learning strategies~\cite{Zhang_2018_CVPR_progressive,deng2018r3net,Wang_2019_CVPR_iterative,wei2019f3net}, attention mechanisms~\cite{liu2018picanet,Chen_2018_ECCV_ras,Wu_2019_CVPR, tian2020weakly, tian2021learning}, and edge/boundary cues~\cite{Qin_2019_CVPR_basnet,Liu_2019_CVPR,Zhao_2019_ICCV_egnet}. \final{Some context-aware methods \cite{wang2019inferring, wang2019salient} are also discussed in SOD fields.} Since glass regions and the content behind the glass surface are not always salient, SOD methods, however, cannot directly address glass segmentation due to a lack of salient features.}
\newylt{Giving a complete review on SOD is beyond the scope of this work. We refer readers to recent survey and benchmark papers \cite{wang2021salient, fan2021salient} for more details.}

\subsection{Specific Region Segmentation}
% \mhy{Specific Region Segmentation (SRS) we defined here refers to segmenting the specific region such as shadow \cite{Hu_2018_CVPR_dsc,Le_2018_ECCV_ad_net,Zhu_2018_ECCV_bdrar,zheng2019distraction}, water \cite{Han_2018_ECCV_water}, mirror \cite{yang2019mirrornet,Mei_2021_CVPR_Depth}, and transparency \cite{Mei_2020_CVPR,xie2020segmenting,xie2021segmenting} in the scene. Such regions are special and have a critical impact on the vision systems. For the shadow, water, and mirror region, there typically exists intensity or content discontinuities between the target region and background. Instead, both the intensity and content are typically similar between the glass region and the background, leading to a great challenge of glass segmentation. Transparency segmentation \cite{xie2020segmenting,xie2021segmenting} considers both stuff such as window, door, and guardrail and things such as cup and eyeglass. We follow \cite{Mei_2020_CVPR} in this paper to focus on the segmentation of stuff (\ie, relatively large glass region that brings great challenges to vision systems), studying the method that is exclusively about grouping glass but not about distinguishing between stuff and things.}

\mhy{Specific Region Segmentation (SRS) we defined here refers to segmenting the specific region such as shadow \cite{Hu_2018_CVPR_dsc,Le_2018_ECCV_ad_net,Zhu_2018_ECCV_bdrar,zheng2019distraction}, water \cite{Han_2018_ECCV_water}, opacity \cite{Yang2018Active}, mirror \cite{yang2019mirrornet,Mei_2021_CVPR_Depth}, and transparency \cite{Mei_2020_CVPR,xie2020segmenting,xie2021segmenting, lin2021rich, He_2021_ICCV} in the scene. Such regions are special and have a critical impact on the vision systems. For the shadow, water, opacity, and mirror region, there typically exists intensity or content discontinuities between the target region and background. Instead, both the intensity and content are typically similar between the glass region and the background, leading to a great challenge of glass segmentation. Transparency segmentation methods \cite{xie2020segmenting,xie2021segmenting, He_2021_ICCV} consider both stuff such as window, door, and guardrail and things such as cup and eyeglass. We follow \cite{Mei_2020_CVPR, lin2021rich} in this paper to focus on the segmentation of stuff (\ie, relatively large glass region that brings great challenges to vision systems), studying the method that is exclusively about grouping glass but not about distinguishing between stuff and things.}

\begin{figure}
	\begin{center}
		\includegraphics[width = 1\linewidth]{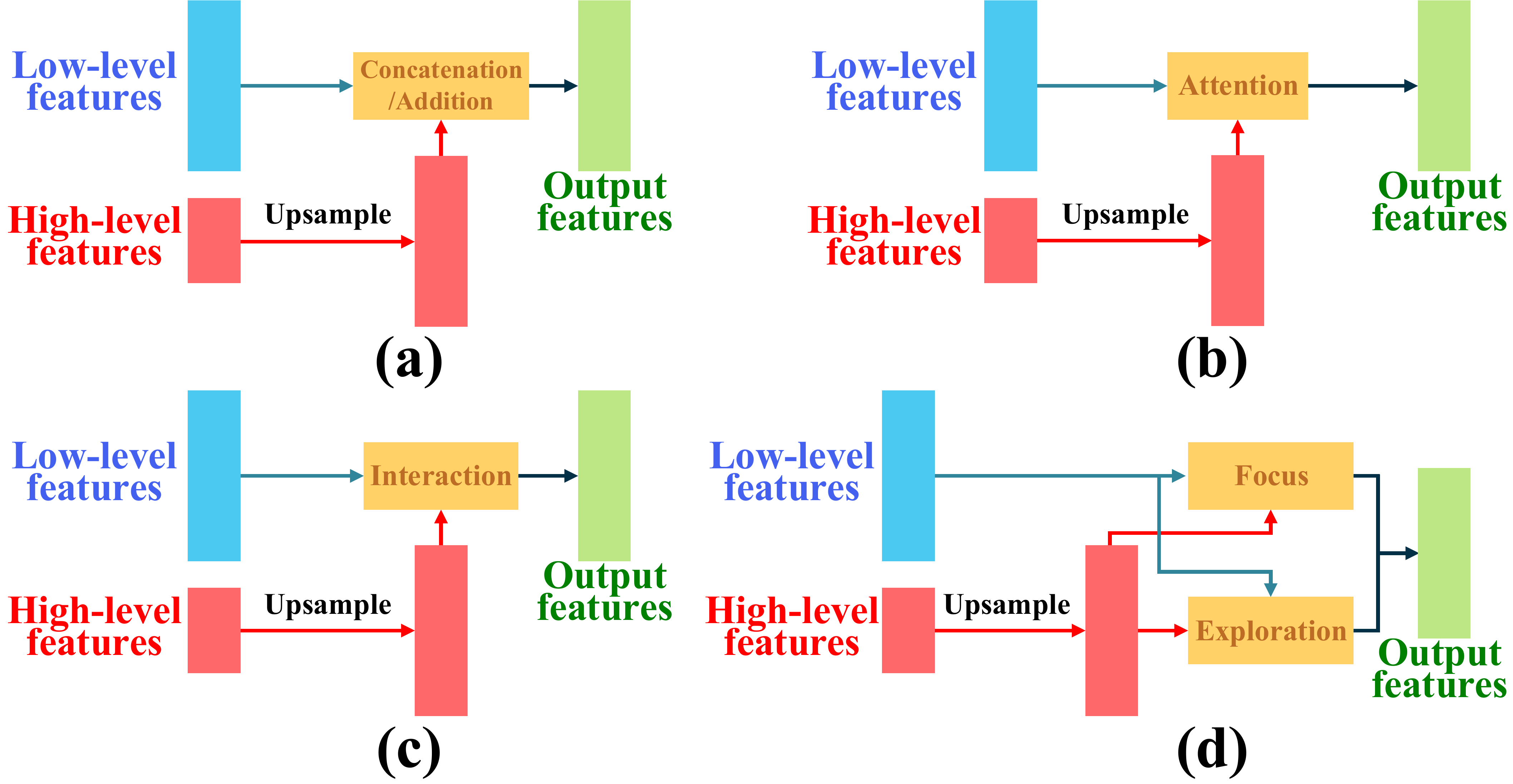}
	\end{center}
	\vspace{-6pt}
	\caption{\final{Different strategies for fusing level-different features. (a) Fusion strategies of channel-wise concatenations or element-wise additions. (b) Fusion strategies use the attention map generated from the higher-level features to guide the current-level features. (c) Fusion strategies of mutual refinement or interaction. (d) Fusion strategies by simultaneously highlighting the common and exploring the difference between level-different features. } }
	\label{fig:fusion}
\end{figure}

% \ylt{Compared with glass regions, mirrors reflect objects in the scene, and the contents presented in the mirror is actually the scene in front of the mirror, these reflection and position cues can help roughly locate the mirrors. For the glass, it has weaker reflectivity, stronger transparency, and more indistinguishability between the glass region and its surroundings. }

% The key difference between glass segmentation and transparent segmentation is that glass segmentation is exclusively about grouping but not about identifying categories. We show that the glass segmentation task plays a vital role in scene understanding and intelligent decision but remains unsolved. Following Mei \etal \cite{Mei_2020_CVPR}, we focus in this paper on the segmentation of large glass region (\eg, glass wall and guardrail) but not small glass-made objects such as cup and wine glass that may also be addressed by the existing segmentation methods.

\begin{figure*}
	\begin{center}
		\includegraphics[width = 1\linewidth]{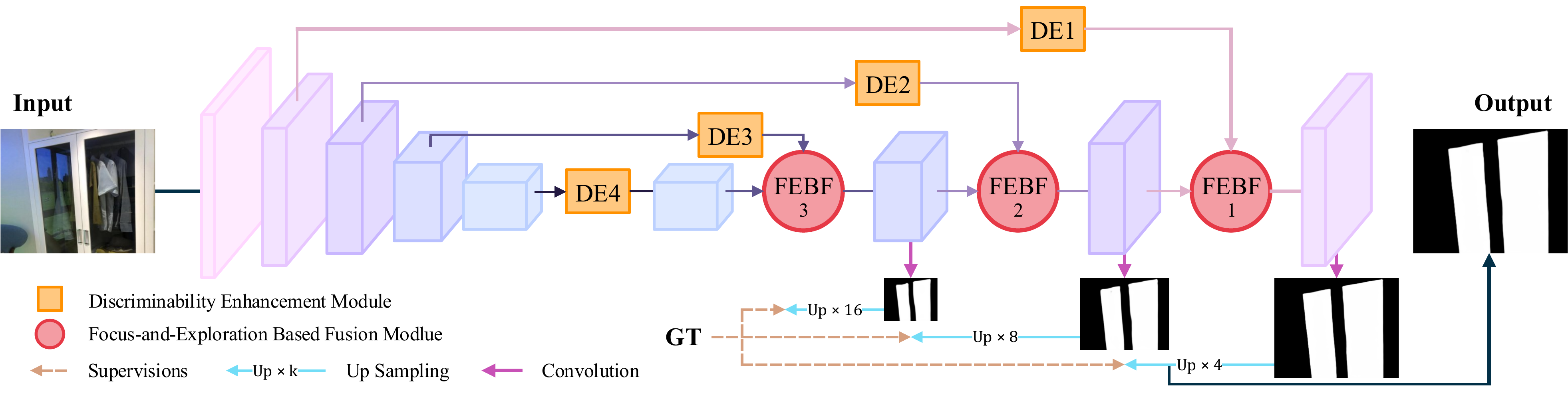}
	\end{center}
	\caption{\mhy{The architecture of our proposed progressive glass segmentation network (PGSNet).}}
	\label{fig:pipeline}
\end{figure*}

\subsection{Deep Feature Fusion}
\mhy{Deep Feature Fusion plays an important role in achieving high performance for many computer vision tasks.}
% \ylt{Both low level tasks\cite{yang2018drfn, Zhang_2020_ICME} and high level tasks\cite{Mei_2021_TCSVT, Mei_2021_CVPR_camouflaged, Mei_2021_CVPR_Depth} utilize deep feature fusion methods to achieve better performances.}
\mhy{The common practice of fusing level-different features is to first upsample higher-level features via interpolation or transposed convolution and then perform channel-wise concatenation \cite{Zhang_2017_ICCV,yang2018drfn,HouPami19Dss,Zhang_2020_ICME,zhang2021two,Qin_2019_CVPR_basnet, Qiao2020multi, liu2021tripartite} or element-wise addition \cite{liu2021prior,Mei_2021_CVPR_camouflaged} between upsampled higher-level features and current-level features (Figure \ref{fig:fusion} (a)). Some methods \cite{Wu_2019_CVPR,yang2019mirrornet,Mei_2020_CVPR, Qiao_2020_CVPR} adopt the attention mechanism in the fusion process, \ie, using the attention map generated from the higher-level features to guide the current-level features (Figure \ref{fig:fusion} (b)). The strategies of mutual refinement \cite{wei2019f3net} or interaction \cite{Pang_2020_CVPR_minet} are also developed (Figure \ref{fig:fusion} (c)).
Our method differs from the above fusion strategies by simultaneously highlighting the common and exploring the difference between level-different features (Figure \ref{fig:fusion} (d)). And we validate the effectiveness of such design by the experiments.}

% \ylt{Describing the current fusion strategies, and then detailing the differences between these method and our fusion strategy, \ie, highlighting the novelty of our focus and exploration fusion stategy.}
% \ylt{The easiest fusion strategy is briefly upsampling the high-level features and pixel-wise summed or concatenated with low-level features. In recent years, the architectures of these deep feature fusion methods are usually based on feature interaction and }

\section{Methodology}
\subsection{Motivation}
To identify the existence of glass, human beings typically first \textbf{\textit{explore}} and then \textbf{\textit{combine}} different useful information including both low-level cues (\eg, color difference between inside and outside the glass, blur/bright spot/ghost caused by reflection) and high-level contexts (\eg, semantic relations between different objects). Existing glass segmentation methods make the attempt to explore useful cues from different levels of features in the deep network via large-field contextual feature extraction \cite{Mei_2020_CVPR}, boundary attention \cite{xie2020segmenting}, or global relation modeling implemented by transformer \cite{xie2021segmenting}. However, how to effectively combine the obtained cues/features has rarely been focused and studied in the glass segmentation field. As different cues/features typically vary in level and characteristic, designing a reasonable fusion strategy rather than fusing them naively would naturally benefit more for the accurate glass segmentation.

\mhy{Based on such observation, we focus in this paper on the effective feature fusion towards accurate glass segmentation. To this end, we first develop a novel level-different features fusion strategy which first uses a discriminability enhancement (DE) module (Sec. \ref{sec:de}) to bridge the characteristic gap between different levels of features and then fuses the enhanced features via a simple yet design-intuitive focus-and-exploration based fusion (FEBF) module (Sec. \ref{sec:febf}). We then construct a progressive glass segmentation network (PGSNet) (Sec. \ref{sec:overview}) by embedding our fusion strategy into an encoder-decoder framework, implementing a coarse-to-fine glass segmentation.}

% However, shallow features are low-level for the edges, lines, and corners, while deep features are high-level for measuring object characteristics, classification, and scene parsing. We refer this kind of difference of characteristic granularity as characteristic gap in this paper. This gap exists to disturb the multi-level feature fusion but has rarely been studied.

% \ylt{We observed that, humans recognize the existence of glass well by both concentrate on the low-level cues(e.g. color/texture difference between inside and outside the glass, bright spot/artifacts caused by reflection and refraction) as well as high-level clues(relations among different objects). When it comes to segment the glass region more detailed, humans will focus on the boundary regions or the specious regions to discriminate whether these regions belong to glass, and explore very subtle contextual difference to judge. } 

% \ylt{To this end, first, we leverage the Large-field Contextual Feature Integration (LCFI) block to extract abundant contextual features from a large receptive field for better glass segmentation and feature inferences, and design the LCFI module based on LCFI blocks. Second, we propose our Focus and Exploration Fusion Module to fully extract and fuse the useful local information from different levels. This fusion strategy both values in cross-level feature consistencies and redundancies.}

\subsection{Overview}\label{sec:overview}
\mhy{Figure \ref{fig:pipeline} shows the overview of our proposed progressive glass segmentation network (PGSNet). Given a single RGB image, we first feed it into a ResNeXt-101 \cite{xie2017aggregated} backbone to extract multi-level features which are then fed into four discriminability enhancement (DE) modules to learn more discriminative feature representations for bridging the characteristic gap between level-different features. Second, we leverage the well-designed focus-and-exploration based fusion (FEBF) module to progressively aggregate level-adjacent features from high-level to low-level. Third, we apply a convolution layer with $3\times3$ kernel on the features generated from each of three FEBF modules to predict different levels of the glass segmentation map. Finally, we upsample the prediction map with the largest spatial size to obtain the original image resolution as the output.}

\begin{figure}[tbp]
	\begin{center}
		\includegraphics[width = 1\linewidth]{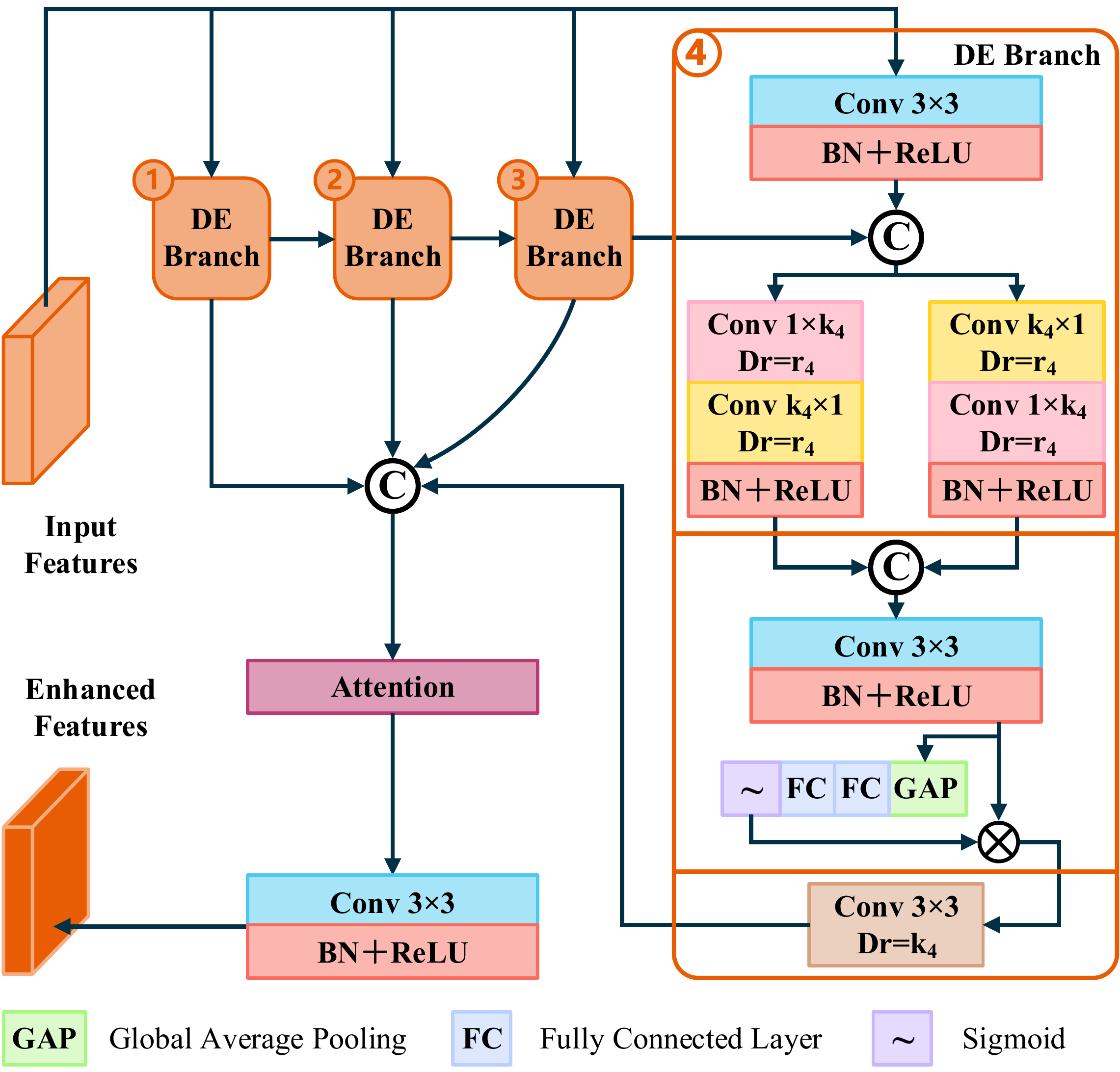}
	\end{center}
	\caption{\mhy{Detailed illustration of our proposed discriminability enhancement (DE) module.}}
	\label{fig:de}
\end{figure}

\begin{figure*}
	\begin{center}
		\includegraphics[width = 1\linewidth]{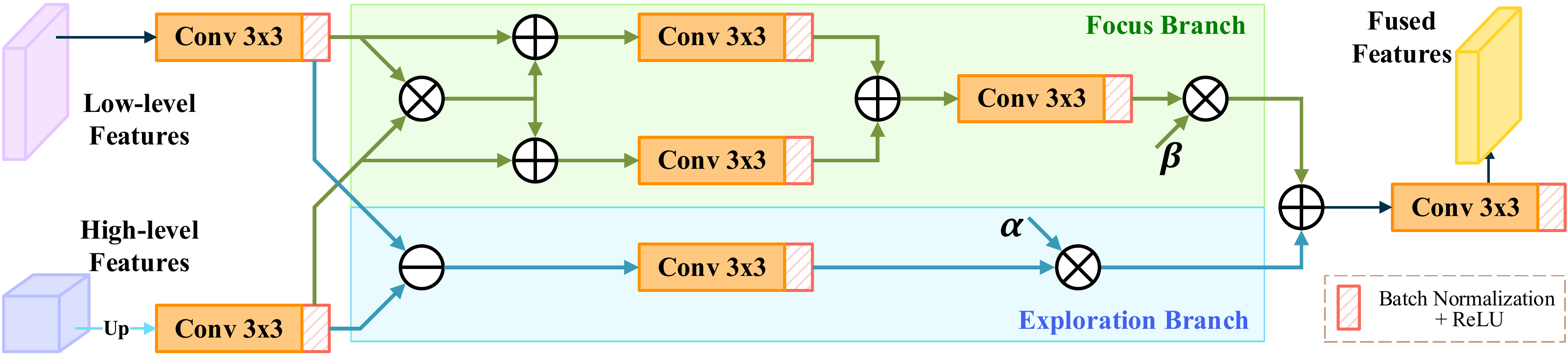}
	\end{center}
	\caption{\mhy{Detailed illustration of our proposed focus-and-exploration based fusion (FEBF) module.}}
	\label{fig:febf}
\end{figure*}

\subsection{Discriminability Enhancement Module}\label{sec:de}
\mhy{Different levels of features typically vary in characteristic. High-level features usually embed more semantic knowledge and contextual relations while low-level features have larger spatial sizes and keep richer and more detailed information such as edges, lines, and corners. The DE module is designed to bridge the characteristic gap by enhancing the discriminability for both high-level and low-level features before fusing them together.}

\mhy{Neuroscience experiments have verified that a set of various sized population receptive fields in the human visual system helps to highlight the area close to the retinal fovea, which is sensitive to small spatial shifts \cite{Liu_2018_ECCV_RFBNet}. This motivates us to use multi-field processing to enhance the discriminability of high-level features for more precise target object location. On the other hand, it has been pointed out in \cite{pang2019towards} that the contexts in a large receptive field could help to enhance the semantics of low-level features, and thus could benefit the suppression of noise and redundant features.
% in the low-level features.
%
These two aspects of analysis inspire us to design the DE module which is applicable for both high-level and low-level features into a multi-branch structure, in which each branch captures different scales of large-field contexts.}

% purify the features to be the more discriminative representations.

% \mhy{In the deep learning era, segmentation methods typically first use deep backbone to harvests features of different levels and then leverage the obtained features to segment the target object/region. Under such framework, we observe that directly using features generated from the backbone to segment the glass would lead to sub-optimal results as not all responses in the features would contribute useful information. Thus a accurate glass segmentation method requires the refinement processing which makes the features more discriminative. Neuroscience experiments have verified that a set of various sized population receptive fields in the human visual system helps to highlight the area close to the retinal fovea, which is sensitive to small spatial shifts [61]. This motivates us to use the multi-field processing to purify the features to be the more discriminative representations.}

\mhy{Figure \ref{fig:de} illustrates the detailed structure of the discriminability enhancement (DE) module. Given the input level-specific features, the DM module aims to purify the features to be the more discriminative representations. Specifically, DE module consists of four parallel DE branches to perform the multi-field processing and the outputs of these four branches are combined to generate the output of the DE module. In each DE branch, the processing can be divided into three steps: local feature extraction (LFE), local feature fusion (LFF), and contextual feature perception (CFP). First, we apply a convolution layer with a kernel size of $k = 3$ for channel reduction and use two parallel $\{1 \times k, k \times 1\}$ and $\{k \times 1, 1 \times k\}$ spatially separable convolutions with the dilation rate of $r$ to efficiently capture the local region information. \final{Practically, for the spatially separable convolutions in four DE branches, the kernel size $k$ is set to 3, 5, 7, 9, and the dilation rate $r$ is set to 1, 2, 3, 4, respectively. } Second, we concatenate the local features in channel and apply a $3 \times 3$ convolution followed by the channel recalibration \cite{hu2018squeeze} for local feature fusion. Third, we employ a $3 \times 3$ convolution with a dilation rate of $r = k$ to perceive contextual information in a large receptive field. Following \cite{Mei_2020_CVPR}, we also add information flow between adjacent branches to facilitate the feature extraction in a larger field-of-view.}

\subsection{Focus-and-Exploration Based Fusion Module}\label{sec:febf}
\mhy{Aggregating different levels of features usually introduces ambiguous features or leads to loss of details, which both make the network fail to optimize. The focus-and-exploration based fusion (FEBF) module is designed to alleviate these two cases via highlighting the common (focus) and exploring the difference (exploration) between level-different features, respectively.}

% \mhy{It is well known that there exists a gap between different levels of features, \ie, deep layer features in the network usually contain more high-level semantic knowledge and are better at locating the exact places of target objects while shallow layer features have larger spatial sizes and keep richer and more detailed low-level information. Leveraging/Fusing both deep and shallow layer features is essential for accurate target object segmentation. The concatenation operation in channel dimension followed by the convolution layer is a common practice to fuse features of different levels. We argue that such naive fusion strategy do not consider the different nature of features of different level and thus would lead to the sub-optimal results. To better bridge the gap between features of different levels, we develop a novel focus-and-exploration based fusion (FBF) module, which reasonably combine different levels of features for better inheriting the informative nature of input features and thus could further benefit the more accurate glass segmentation.}

\mhy{As depicted in Figure \ref{fig:febf}, the FEBF module takes two level-different features as input and outputs the fused features. Specifically, given the input high-level features $F_h \in \mathbb{R}^{C_h \times H_h \times W_h}$ and low-level features $F_l \in \mathbb{R}^{C_l \times H_l \times W_l}$, where $C_{h/l}$, $H_{h/l}$, and $W_{h/l}$ represent the channel number, height, and width, respectively, we first do the following to make them keep the same in spatial size and channel:
% we first upsample the $F_l$ via bilinear interpolation and then apply the convolution followed by batch normalization (BN) and ReLU activation function on both high-level features and upsampled low-level features for the consistency in spatial size and channel:
\begin{align}
F'_h &= \mathcal{R}(\mathcal{N}(\psi_{3\times 3}(\mathcal{U}(F_h)))), \\
F'_l &= \mathcal{R}(\mathcal{N}(\psi_{3\times 3}(F_l))),
\end{align}
where $\psi_{3\times 3}$ is a convolution layer with a kernel size of 3; $\mathcal{U}$ is the bilinear upsampling; $\mathcal{N}$ is the batch normalization (BN); and $\mathcal{R}$ is the ReLU activation function. Then, we highlight the common and explore the difference between $F'_h$ and $F'_l$ in the focus and exploration branch, respectively.
\begin{equation}
\begin{aligned}
F_f = \mathcal{R}(\mathcal{N}(\psi_{3\times 3}(&\mathcal{R}(\mathcal{N}(\psi_{3\times 3}((F'_l \otimes F'_h) \oplus F'_l)))\oplus\\
&\mathcal{R}(\mathcal{N}(\psi_{3\times 3}((F'_l \otimes F'_h) \oplus F'_h)))))),
\end{aligned}
\end{equation}
\begin{equation}
F_e = \mathcal{R}(\mathcal{N}(\psi_{3\times 3}( F'_l \ominus F'_h ))),
\end{equation}
where $\otimes$, $\oplus$, and $\ominus$ denote the element-wise multiplication, addition, and subtraction, respectively.
Finally, we can obtain the output features $F_o \in \mathbb{R}^{C_l \times H_l \times W_l}$ by:
\begin{equation}
F_o = \mathcal{R}(\mathcal{N}(\psi_{3\times 3}(\alpha F_f \oplus \beta F_e))),
\end{equation}
where $\alpha$ and $\beta$ are the learnable scale parameters which are initialized as 1.}
% \newylt{In the focus module, we can view this operation as highlighting the similar regions between level-different features. In the exploration branch, the low-level features and high-level features will be activated after the previous DE module and the features of glass region tend to be white. After the subtraction operation, the remaining uncertain white regions will be further judged whether they are glass regions. The exploration module will find the differences between low-level and high-level features, alleviate both the introduction of ambiguous and loss of details, suppress the redundancy information and enhance the subtle features. For the glass segmentation task, the focus module aims to focus on the coarse large regions and deterministic features, such as plain glass surfaces, while the exploration module aims to excavate fine scattered regions and controversial features, like glass regions with strong reflections, near boundaries or overlapped by objects with complex shape. 
%  The features extracted from the exploration module are then fused with features from focus module, which bring out the final features extracted from FEBF module.
% }

\newylt{Note that the subtraction operation used in our FEBF module differs from \cite{ding2018context} and \cite{li2020improving} in both motivation and implementation. Ding \textit{et al.} \cite{ding2018context} conduct the subtraction between the local and contextual features to perceive the contextual contrasted features for locating discriminative objects. Li \textit{et al.} \cite{li2020improving} perform the subtraction between segmentation features and body features to harvest edge features which are supervised by ground truth edge map for explicitly modeling the different parts of objects. Our motivation to use subtraction operation between different features is to screen out the uncertain areas for further supporting the effective fusion strategy. Different levels of features typically generate different responses related to the glass. For example, high-level features tend to be activated on the coarse location of the glass region while low-level features usually have a high response on the detailed boundaries. The regions with high values in both high-level and low-level features are more likely to belong to the glass region. Thus we use \textit{addition} operation to make the network \textit{focus} on such regions. For the regions with high responses in one features while have low values in another one features are the uncertain areas. We adopt the \textit{subtraction} operation to screen out such regions and then use a convolution layer to further \textit{explore} the true glass regions. Finally, the \textit{focus} features and \textit{exploration} features are combined, generating more precise features for glass segmentation.}

\subsection{Loss Function}
\mhy{We train our PGSNet with a hybrid loss defined as:
\begin{equation}\label{equ:loss_hybrid}
%   \mathcal{L}_{hybrid} = \gamma \ell_{bce} + \lambda \ell_{iou},
\mathcal{L}_{hybrid} = \gamma \mathcal{L}_{bce} + \lambda \mathcal{L}_{iou},
\end{equation}
where $\mathcal{L}_{bce}$ and $\mathcal{L}_{iou}$ denote the binary cross-entropy (BCE) loss \cite{de2005tutorial} and IoU loss \cite{Mattyus_2017_ICCV}, respectively. $\gamma$ and $\lambda$ are the balancing parameters and we empirically set them to 1.
BCE loss \cite{de2005tutorial} is the most widely used loss in binary classification and segmentation tasks, which calculates the loss for each pixel independently. It is a pixel-wise measure which helps with the convergence on all pixels. As BCE loss \cite{de2005tutorial} weights both the foreground and background pixels equally, the loss of foreground pixels will be diluted for images where the background is dominant. Hence, we further include the map-level IoU loss \cite{Mattyus_2017_ICCV} which could make the network focus more on the foreground regions and output the complete segmentation results.
To facilitate the learning process, we adopt the deep supervision \cite{xie2015holistically}. The overall loss function is:
\begin{equation}\label{equ:loss_overall}
\mathcal{L}_{overall} = \sum_{i=1}^{3}2^{(3-i)}\mathcal{L}_{hybrid}^{i},
\end{equation}
where $\mathcal{L}_{hybrid}^{i}$ denotes the hybrid loss between the ground truth glass mask and the prediction from the fused features generated by the FEBF module at the $i$-\textit{th} level.}

% \ylt{We define our loss function ${l}$ as a combination of BCE loss \cite{de2005tutorial} $l_{bce}$ and IOU loss \cite{Qin_2019_CVPR_basnet} $l_{iou}$ to optimize the network during the training process, \ie,  $l =l_{bce} + l_{iou},$ where two parts of $L$ both provide effective pixel-level and image-level supervision for accurate glass segmentation.  The total loss function $l_{all}$ is:$$l_{all} = \sum\limits_{i=1}^{3}w_i l_i, $$where $w_i$ represents the balancing parameters, and $l_i$ is the aformentioned hybrid loss function between the $i$-th upsampled glass map and the ground truth. }

\section{Experiments}
\subsection{Dataset}
\mhy{We evaluate the effectiveness of our method on two benchmark datasets GDD \cite{Mei_2020_CVPR} and Trans10K-Stuff \cite{xie2020segmenting} as well as our newly constructed home-scene-oriented (HSO) glass segmentation dataset. GDD \cite{Mei_2020_CVPR} is the first glass segmentation dataset in the deep learning era, which consists of 2,980 training images and 936 testing images. Trans10K \cite{xie2020segmenting} is a large-scale transparent object segmentation dataset, containing two categories of objects, \ie, stuff and things. As the glass segmentation task we strive to address is exclusively about grouping but not about identifying object categories, we only use the ``stuff' images and corresponding annotations in Trans10K \cite{xie2020segmenting}. In our experiment, 2,455 glass image/mask pairs in Trans10K-Stuff \cite{xie2020segmenting} are used for training and 1,771 images in both validation and testing set are used for testing.}
%
% Unlike these two existing datasets, the goal of our new dataset is to provide more diverse, more challenging, and high quality examples of glass at home, for facilitating further study on this topic and practical applications, especially for household robots.}

\begin{figure*}[t]
	\begin{center}
		\includegraphics[width = 1\linewidth]{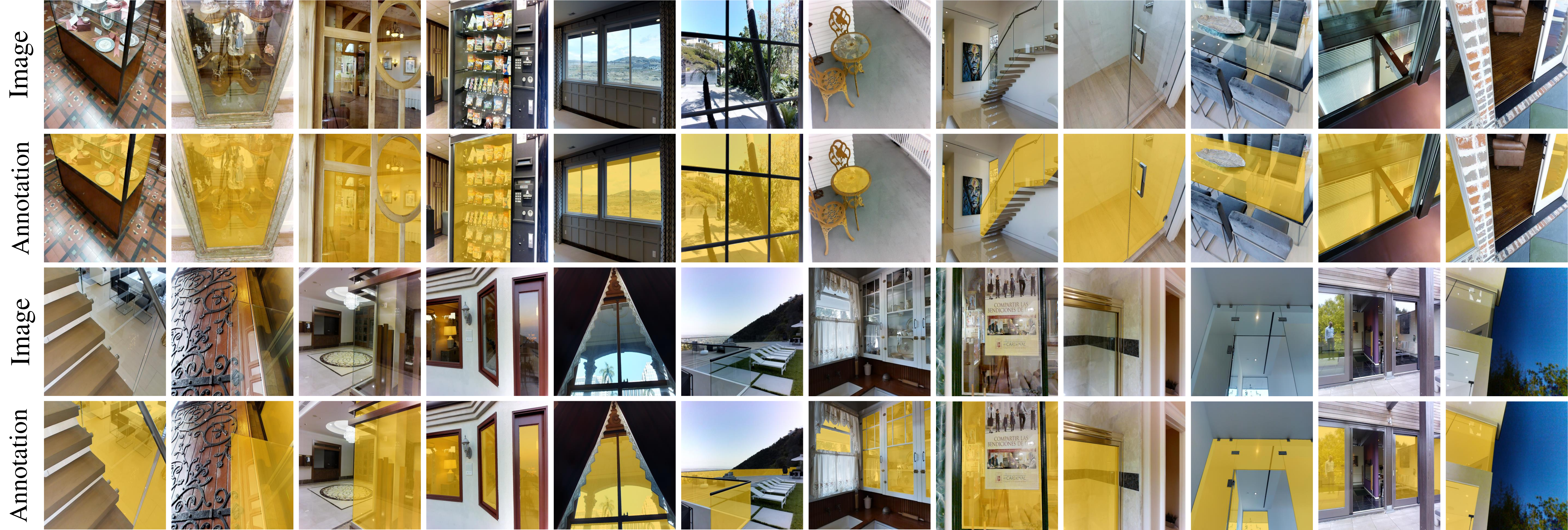}
	\end{center}
	\caption{\mhy{Example glass image/mask pairs in our HSO dataset. It shows that HSO covers diverse types of glass in various home scenes.}}
	\label{fig:example}
\end{figure*}

\mhy{Our HSO dataset is built to further diversify the patterns of glass, especially the glass in home scenes, for advancing household robot applications and in-depth research on this topic.
Instead of capturing the glass images ourselves, we compose the HSO dataset from selected exemplars from four popular datasets (\ie, Matterport3D~\cite{Matterport3D}, SUNRGBD~\cite{song2015sun}, ScanNet~\cite{dai2017scannet}, and 2D3DS~\cite{armeni2017joint}) to ensure a wide diversity and broad coverage. See Table \ref{tab:dataset_composition} for a summary and Figure \ref{fig:example} for representative examples. Each selected image contains at least one glass region, and the pixel-level accurate reference glass masks are created by professional annotators. The statistics of HSO dataset are shown in Figure \ref{fig:dataset}, where we can see that HSO has glass with reasonable property distributions in terms of location, area, scene, and global color contrast. We will release HSO to stimulate further research on this topic.}

\def\wsample{0.5\linewidth}
\def\hsample{1in}
\begin{figure}[tbp]
	\setlength{\tabcolsep}{1.5pt}
	\centering
	\begin{tabular}{cc}
		\includegraphics[width=\wsample, height=\hsample]{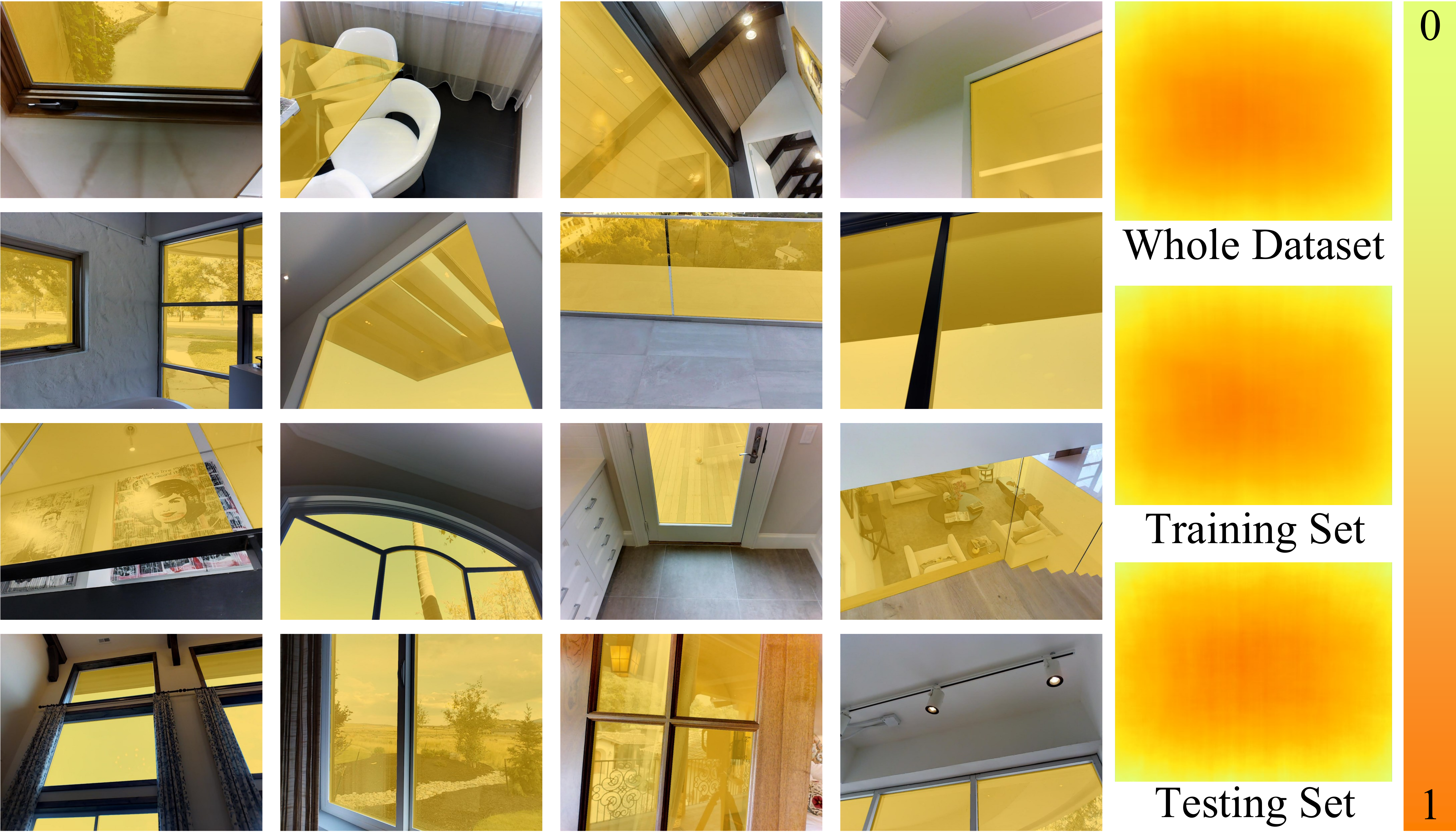}&
		\includegraphics[width=\wsample, height=\hsample]{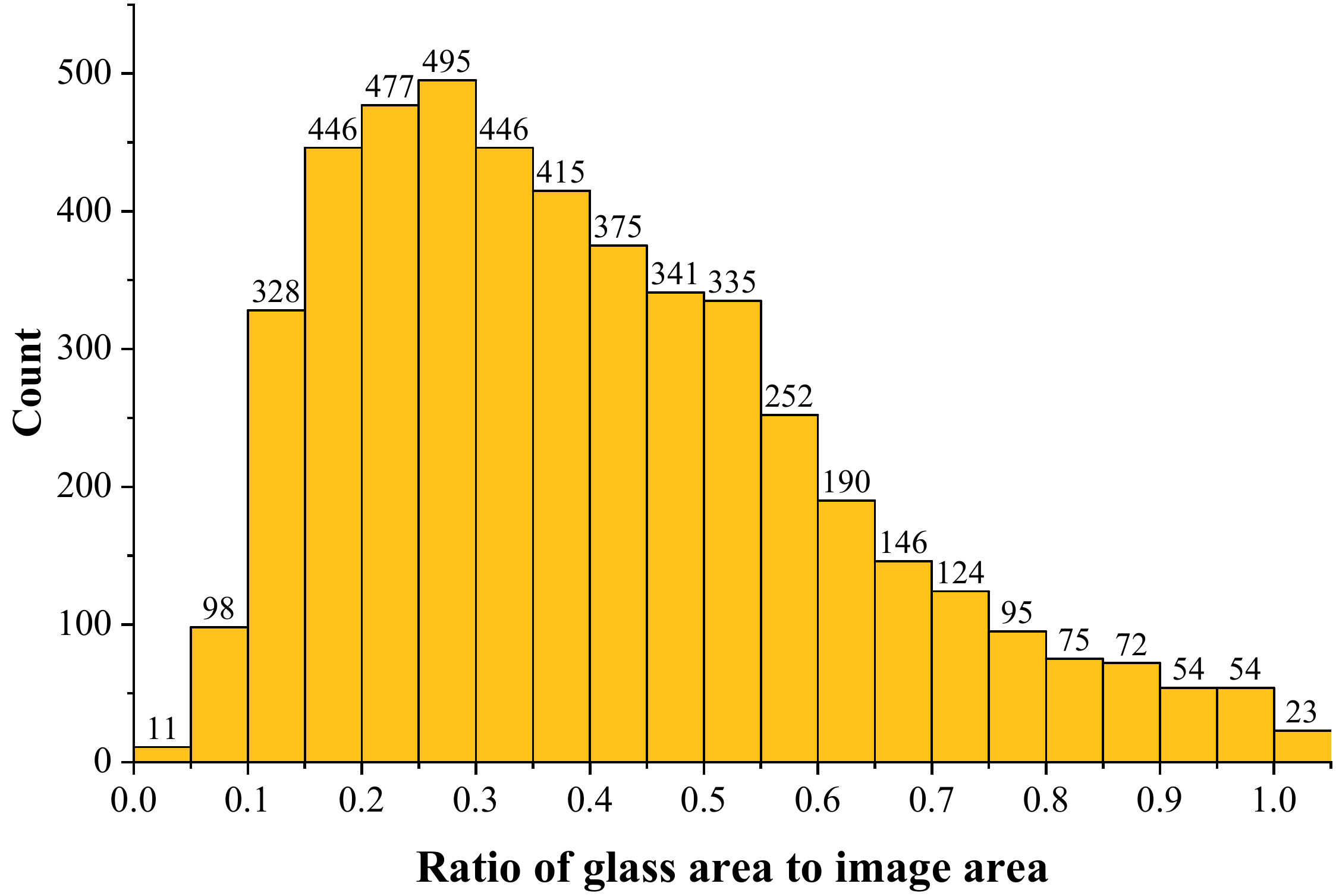} \\
		{\small (a) glass location distribution} & {\small (b) glass area distribution} \vspace{0.1in} \\
		\includegraphics[width=\wsample, height=\hsample]{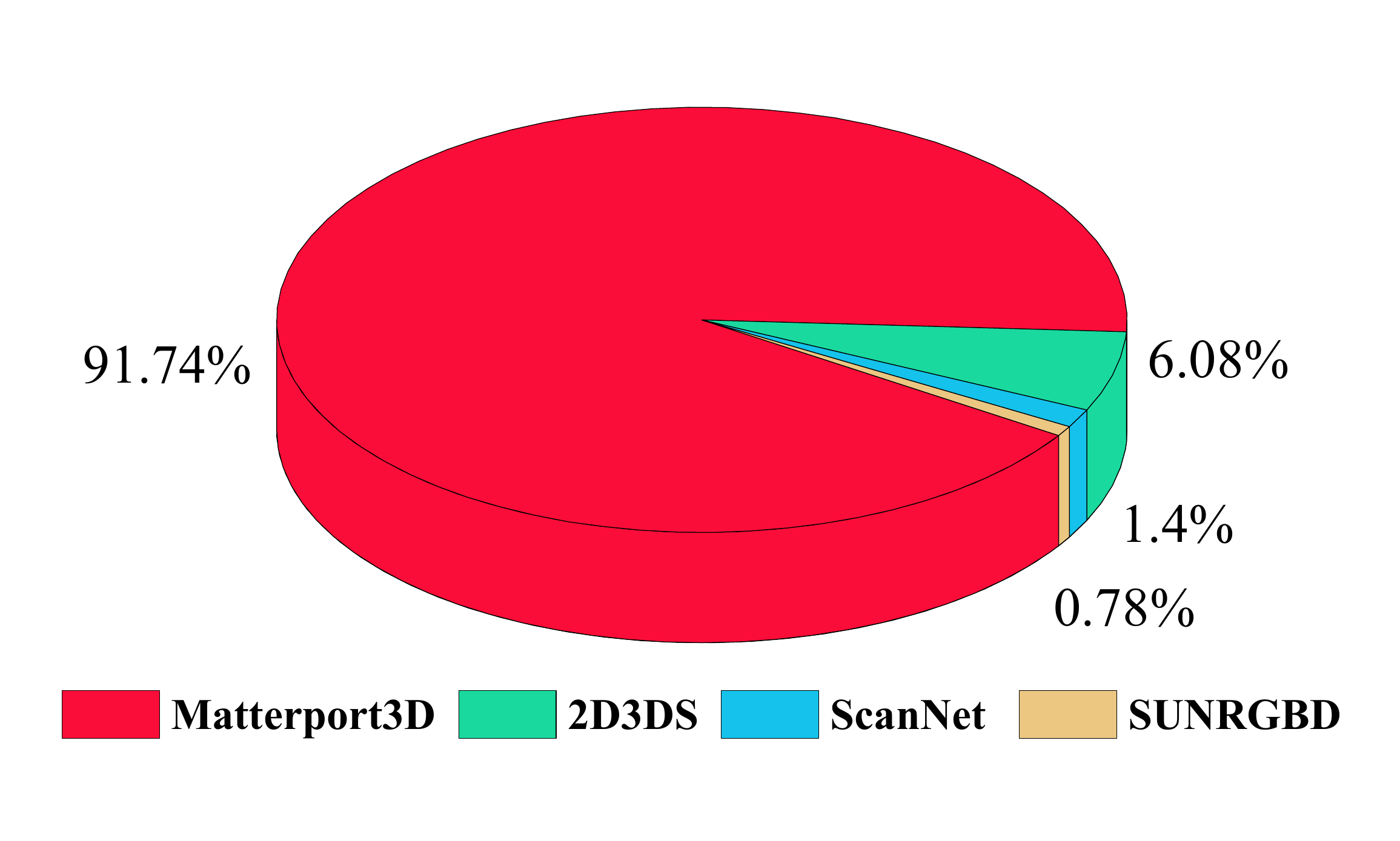}&
		\includegraphics[width=\wsample, height=\hsample]{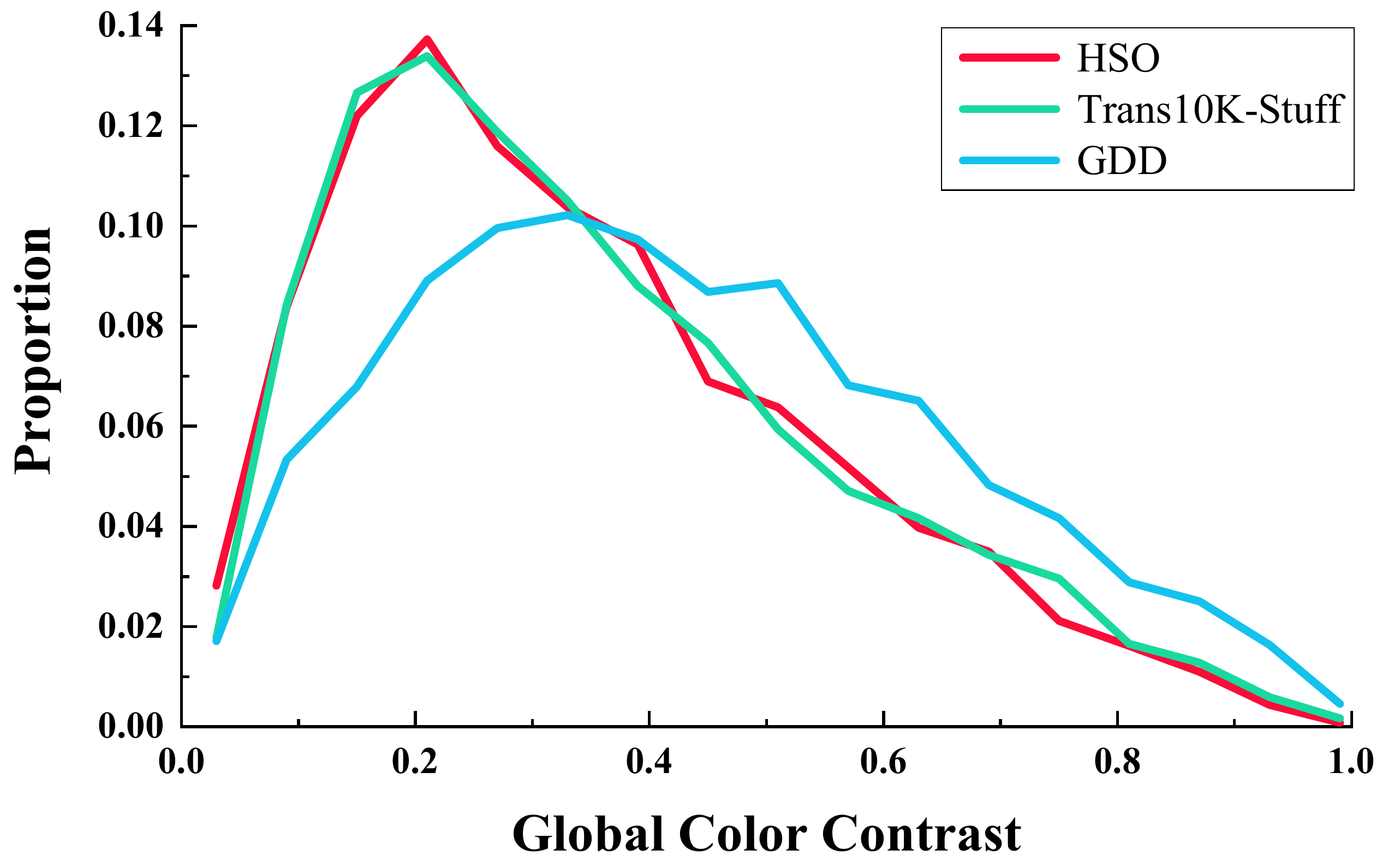} \\
		{\small (c) glass composition distribution} & {\small (d) glass contrast distribution} \vspace{0.1in} \\
	\end{tabular}
	\caption{\mhy{Statistics of our HSO dataset.}}
	\label{fig:dataset}
\end{figure}

\begin{table}[tbp]
	\centering
	\setlength{\tabcolsep}{2.4pt}
	\footnotesize
	\centering
    \caption{\mhy{Composition of our HSO dataset.}}
	\begin{tabular}{p{2.4cm}<{\centering}|p{1.5cm}<{\centering}|p{1.5cm}<{\centering}|p{1.5cm}<{\centering}}
		\hline
		\hline
		Dataset & Images & Train & Test  \\
		\hline
		% 		\hline
		Matterport3D \cite{Matterport3D} & 4,451 & 2,887 & 1,564  \\
		\hline
		2D3DS \cite{armeni2017joint} & 295 & 162 & 133  \\
		\hline
		ScanNet \cite{dai2017scannet} & 68 & 17 & 51  \\
		\hline
		SUNRGBD \cite{song2015sun} & 38 & 4 & 34 \\
		\hline
		\textbf{Total} & \textbf{4,852} & \textbf{3,070} & \textbf{1,782} \\
		\hline
		\hline
	\end{tabular}
	\label{tab:dataset_composition}
\end{table}

\subsection{Experimental Setup}
\subsubsection{Evaluation Metrics}
% \mhy{For a comprehensive evaluation, we adopt four widely used metrics for quantitatively assessing the glass segmentation performance: intersection over union ($IoU$), weighted F-measure ($F_\beta^w$) \cite{margolin2014evaluate_wfmeasure}, mean absolute error ($MAE$), and balance error rate ($BER$) \cite{nguyen2017shadow_ber}.}
% %
\mhy{For a comprehensive evaluation, we adopt four widely used metrics for quantitatively assessing the mirror segmentation performance: intersection over union ($IoU$), weighted F-measure ($F_\beta^w$) \cite{margolin2014evaluate_wfmeasure}, mean absolute error ($MAE$), and balance error rate ($BER$) \cite{nguyen2017shadow_ber}.}

\add{The intersection over union ($IoU$) is widely used in the segmentation field, which is defined as:
\begin{equation}\label{equ:iou}
	IoU = \frac{\sum\limits_{i=1}^{H}\sum\limits_{j=1}^{W}(G(i,j)*P_b(i,j))}{\sum\limits_{i=1}^{H}\sum\limits_{j=1}^{W}(G(i,j)+P_b(i,j)-G(i,j)*P_b(i,j))},
\end{equation}
where $G$ is the ground truth mask in which the values of the glass region are 1 while those of the non-glass region are 0; $P_b$ is the predicted mask binarized with a threshold of 0.5; and $H$ and $W$ are the height and width of the ground truth mask, respectively.}

\add{We also adopt the weighted F-measure metric from the salient object detection field. F-measure ($F_\beta$) is a comprehensive measure on both the precision and recall of the prediction map. Recent studies \cite{fan2017structure_smeasure,fan2018enhanced_emeasure} have suggested that the weighted F-measure ($F_\beta^w$) \cite{margolin2014evaluate_wfmeasure} can provide more reliable evaluation results than the traditional $F_\beta$. Thus, we report $F_\beta^w$ in the comparison.}

\add{The mean absolute error ($MAE$) metric is widely used in foreground-background segmentation tasks, which calculates the element-wise difference between the prediction map $P$ and the ground truth mask $G$:
\begin{equation}\label{equ:mae}
	MAE = \frac{1}{H\times W}\sum_{i=1}^{H}\sum_{j=1}^{W}|P(i,j)-G(i,j)|,
\end{equation}
where $P(i,j)$ indicates the predicted probability score at location $(i,j)$.}

\add{The last metric is the balance error rate ($BER$), which is a standard metric in the shadow detection field, defined as:
\begin{equation}\label{equ:ber}
	BER = (1 - \frac{1}{2}(\frac{TP}{N_{p}} + \frac{TN}{N_{n}})) \times 100,
\end{equation}
where $TP$, $TN$, $N_p$, and $N_n$ represent the numbers of true positive pixels, true negative pixels, glass pixels, and non-glass pixels, respectively.}

\mhy{Note that for $IoU$ and $F_\beta^w$, it is the higher the better, while for $MAE$ and $BER$, it is the lower the better.}

\subsubsection{Implementation Details}
\mhy{We implement our model with the PyTorch toolbox \cite{paszke2019pytorch}. We train and test our model on a PC with an Intel Core i7-7700K 4.2GHz CPU (with 32GB RAM) and an NVIDIA Titan V GPU (with 12GB memory). The strategies of training and testing on three datasets keep the same. Specifically, for training, input images are augmented by randomly horizontal flipping and resizing. The parameters of the backbone network are initialized with the ResNeXt-101 model \cite{xie2017aggregated}  pre-trained on ImageNet \cite{deng2009imagenet} while the remaining layers of our model are initialized randomly. We use the stochastic gradient descent (SGD) optimizer with the momentum of 0.9 and the weight decay of $5\times10^{-4}$ for loss optimization. We set the batch size to 6 and adjust the learning rate by the poly strategy \cite{liu2015parsenet_poly} with the basic learning rate of 0.001 and the power of 0.9. It takes about 14, 16, and 21 hours for the network to converge after 170, 220, and 250 epochs for the training on GDD \cite{Mei_2020_CVPR}, Trans10K-Stuff \cite{xie2020segmenting}, and our HSO dataset, respectively. For testing, the image is first resized to $352\times352$ for network inference and then the output map is resized back to the original size of the input image. Both the resizing processes use bilinear interpolation. We do not use any post-processing such as the fully connected conditional random field (CRF) \cite{krahenbuhl2011efficient} to further enhance the final output. The inference for a $352\times352$ image takes only 0.047 seconds (about 21 FPS).}

% \begin{table}[htbp]
% 	\centering
% 	\setlength{\tabcolsep}{2.4pt}
% 	\footnotesize
% 	\centering
% 	\begin{tabular}{p{2.0cm}<{\centering}|p{1.2cm}<{\centering}|p{1.2cm}<{\centering}|p{1.2cm}<{\centering}|p{1.2cm}<{\centering}|p{1.2cm}<{\centering}}
% 		\hline
% 		\hline
% 		Methods & Pub.' Year & Input Sizes & FLOPs (G) & Params (M) & Model Size (MB) \\
% 		\hline
% 		\hline
% 		BASNet \cite{Qin_2019_CVPR_basnet} & CVPR'19 & 256x256 & 127.444 & 87.060 & - \\
% 		PoolNet \cite{Liu_2019_CVPR} & CVPR'19 & 320x320  & 76.843 & 68.261 & -  \\
% 		MINet-R \cite{Pang_2020_CVPR_minet} & CVPR'20 & 320x320 & 87.032 & 162.378 & - \\
% 		ITSD \cite{Zhou_2020_CVPR_ITSD} & CVPR'20 & 320x320 & 15.94 & 26.07 & 102 \\
% 		\hline
% 		\hline
% 		MirrorNet$^\dag$ \cite{yang2019mirrornet} & ICCV'19 & 384x384 & 77.656 & 121.767 & 466 \\
% 		\hline
% 		GDNet \cite{Mei_2020_CVPR} & CVPR'20 & 416x416 & 271.533 & 201.720 & 771 \\
% 		TransLab \cite{xie2020segmenting} & ECCV'20 & 512x512 & 61.264 & 40.147 & 169    \\
% 		Trans2Seg \cite{xie2021segmenting} & arXiv'21 & 512x512 & 49.034 & 56.215 & 703 \\
% 		\hline
% 		Ours & Submit'21 & 352x352 & 76.210 & 184.872 & 707 \\
% 		\hline
% 	\end{tabular}
% 	\caption{Comparison of the computational efficiency of different methods. For each method, we list FLOPs (G), number of parameters (M), model size (MB), \ylt{and inference time (ms)}. For MirrorNet~\cite{yang2019mirrornet}, we report the CRF~\cite{krahenbuhl2011efficient} post-processing time in {\color{cyan} Cyan}.}
% 	\label{tab:comparison_flops}
% \end{table}

\begin{table*}[htbp]
	\centering
	\setlength{\tabcolsep}{2.8pt}
	% 	\footnotesize
	\small
	\centering
	\caption{\mhy{Quantitative comparison to the state-of-the-art methods on the GDD \cite{Mei_2020_CVPR}, Trans10K-Stuff \cite{xie2020segmenting} and our newly constructed HSO dataset. All the methods are re-trained on the corresponding training set. $\dag$ denotes using CRFs~\cite{krahenbuhl2011efficient} for post-processing. ``Statistics" means thresholding glass location statistics from the corresponding training set as a glass mask for segmentation. $\bullet$: semantic segmentation method. $\circ$: salient object detection method. $\vartriangle$: shadow detection methods. $\S$: medical image segmentation method. *: mirror segmentation method. $ \diamond $: transparent object segmentation. $\star$: glass segmentation methods. The first, second, and third best results are marked in \red{\textbf{red}}, \green{\textbf{green}}, and \blue{\textbf{blue}}, respectively. Our method achieves the best performance on all three challenging datasets under four standard metrics.}}
	\begin{tabular}{p{2.41cm}|p{1.545cm}<{\centering}|p{1.745cm}<{\centering}|p{0.75cm}<{\centering}|p{0.75cm}<{\centering}|p{0.75cm}<{\centering}|p{0.75cm}<{\centering}|p{0.75cm}<{\centering}|p{0.75cm}<{\centering}|p{0.75cm}<{\centering}|p{0.75cm}<{\centering}|p{0.75cm}<{\centering}|p{0.75cm}<{\centering}|p{0.75cm}<{\centering}|p{0.75cm}<{\centering}}
		\hline
		\hline
		\multirow{3}{*}{Methods} & \multirow{3}{*}{Pub.'Year} & \multirow{3}{*}{Backbone} & \multicolumn{4}{c|}{GDD  \cite{Mei_2020_CVPR} } & \multicolumn{4}{c|}{Trans10K-Stuff \cite{xie2020segmenting}} & \multicolumn{4}{c}{HSO (Ours)}\\
		\cline{4-15}
		&  & & \multicolumn{2}{c}{Trainset:2,980} & \multicolumn{2}{c|}{ Testset:936 } & \multicolumn{2}{c}{Trainset:2,455} &
		\multicolumn{2}{c|}{Testset:1,771 } &
		\multicolumn{2}{c}{Trainset:3,070 } &
		\multicolumn{2}{c}{Testset:1,782} \\
		\cline{4-15}
		&  & & IoU$\uparrow$ & $F_\beta^w$$\uparrow$ & MAE$\downarrow$ & BER$\downarrow$ & IoU$\uparrow$ & $F_\beta^w$$\uparrow$ & MAE$\downarrow$ & BER$\downarrow$ & IoU$\uparrow$ & $F_\beta^w$$\uparrow$ & MAE$\downarrow$ & BER$\downarrow$\\
		\hline
		Statistics & - & - & 40.75 & 0.427 & 0.451 & 39.31 & 45.88 & 0.448 & 0.371 & 27.28 & 28.86 & 0.311 & 0.485 & 44.84 \\
		\hline
		ICNet$^\bullet$ \cite{Zhao_2018_ECCV_icnet} & ECCV'18 & ResNet-50  & 69.59 & 0.747 & 0.164 & 16.10 & 74.94 & 0.784 & 0.110 & 10.92 & 62.15 & 0.674 & 0.165 & 17.07 \\
		PSPNet$^\bullet$ \cite{zhao2017pyramid} & CVPR'17 & ResNet-50 & 84.06 & 0.867 & 0.084 & 8.79 & 87.89 & \textbf{\blue{0.907}} & \textbf{\blue{0.045}} & 5.46 & 77.60 & 0.814 & 0.095 & 10.57\\
		DeepLabv3+$^\bullet$ \cite{chen2018encoder} & ECCV'18 & ResNet-50 & 69.95 & 0.767 & 0.147 & 15.49 & 51.52 & 0.602 & 0.229 & 23.80 & 64.47 & 0.705 & 0.149 & 16.03\\
		DenseASPP$^\bullet$ \cite{Yang_2018_CVPR_DenseASPP} & CVPR'18 & ResNet-50  & 83.68 & 0.867 & 0.081 & 8.66 & 86.34 & 0.894 & 0.051 & 6.12 & 75.94 & 0.805 & 0.096 & 11.34  \\
		BiSeNet$^\bullet$ \cite{Yu_2018_ECCV_BiSeNet} & ECCV'18 & ResNet-50 & 80.00 & 0.830 & 0.106 & 11.04 & 85.82 & 0.885 & 0.056 & 6.11 & 75.85 & 0.798 & 0.101 & 11.04  \\
		% 		PSANet$^\bullet$ \cite{Zhao_2018_ECCV_psanet} & ECCV'18 & 83.52 & 0.862 & 0.082 & 9.09 &&&&& 78.21 & 0.823 & 0.088 & 10.23  \\
		DANet$^\bullet$ \cite{Fu_2019_CVPR_DANet} & CVPR'19 & ResNet-50 & 84.15 & 0.864 & 0.089 & 8.96 & 88.18 & \textbf{\blue{0.907}} & \textbf{\blue{0.045}} & 5.28 & 77.69 & 0.817 & \textbf{\green{0.091}} & 10.60  \\
		CCNet$^\bullet$ \cite{Huang_2019_ICCV_CCNet} & ICCV'19 & ResNet-50 & 84.29 & 0.867 & 0.085 & 8.63 &88.20 & 0.906 & \textbf{\green{0.044}} & 5.15 & 78.17 & \textbf{\blue{0.820}} & \textbf{\blue{0.092}} & 10.34  \\
		GFFNet$^\bullet$ \cite{li2020gated} & AAAI'20 & ResNet-50 & 82.41 & 0.855 & 0.090 & 9.11 & 69.29 & 0.747 & 0.143 & 14.19 & 77.34 & 0.810 & 0.094 & 9.69 \\
		SFNet$^\bullet$ \cite{li2020semantic} & ECCV'20 & ResNet-50 & 80.96 & 0.848 & 0.102 & 10.23 & 71.27 & 0.767 & 0.133 & 13.14 & 77.48 & 0.814 & \textbf{\green{0.091}} & 10.79\\		
		FaPN$^\bullet$ \cite{huang2021fapn} & ICCV'21 & ResNet-101 & 86.65 & 0.887 & \textbf{\red{0.062}} & \textbf{\blue{5.69}} & \textbf{\blue{89.09}} & \textbf{\green{0.913}} & \textbf{\red{0.042}} & 4.80 & 78.05 & \textbf{\green{0.835}} & \textbf{\red{0.089}} & \textbf{\blue{9.51}} \\
		\hline
		DSS$^\circ$ \cite{HouPami19Dss} & TPAMI'19 & ResNet-50   & 80.24 & 0.799 & 0.123 & 9.73 & 84.77 & 0.855 & 0.075 & 6.42 & 73.08 & 0.730 & 0.135 &  12.04 \\
		PiCANet$^\circ$ \cite{liu2018picanet} & CVPR'18 & ResNet-50  & 83.74 & 0.848 & 0.093 & 8.24 & 83.99 & 0.843 & 0.077 & 7.03 & 71.66 & 0.730 & 0.148 & 13.31  \\
		RAS$^\circ$ \cite{Chen_2018_ECCV_ras} & ECCV'18 & ResNet-50  & 80.96 & 0.830 & 0.106 & 9.48 & 85.40 & 0.882 & 0.062 & 6.20 & 74.63 & 0.775 & 0.116 & 11.24  \\
		% 		R\textsuperscript{3}Net$^{\circ\dag}$ \cite{deng2018r3net} & IJCAI'18 & 76.72 & 0.796 & 0.132 & 13.84 & 87.41 & 0.900 & 0.049 & 5.29 & 78.45 & 0.828 & 0.087 & 9.61 \\
		CPD$^\circ$ \cite{Wu_2019_CVPR} & CVPR'19 & ResNet-50  & 82.52 & 0.850 & 0.095 & 8.87 & 86.08 & 0.869 & 0.064 & 5.89 & 76.16 & 0.789 & 0.111 &  10.58  \\
		% 		PoolNet$^\circ$ \cite{Liu_2019_CVPR} & CVPR'19 & 81.92 & 0.840 & 0.100 & 8.95 &&&&&77.29 & 0.796 & 0.105 & 10.02  \\
		% 		BASNet$^\circ$ \cite{Qin_2019_CVPR_basnet} & CVPR'19 & 82.88 & 0.854 & 0.094 & 8.70 &&&&&&&& \\
		EGNet$^\circ$ \cite{Zhao_2019_ICCV_egnet} & ICCV'19 & ResNet-50 & 85.05 & 0.870 & 0.083 & 7.43 & 84.57 & 0.863 & 0.068 & 6.59 & 74.29 & 0.771 & 0.119 & 11.58 \\
		F3Net$^\circ$ \cite{wei2019f3net}  & AAAI'20 & ResNet-50 & 84.79 & 0.870 & 0.082 & 7.38 & 86.23 &  0.881 & 0.061 & 5.81 & 76.84 & 0.799 & 0.105 & 10.58  \\
		MINet-R$^\circ$ \cite{Pang_2020_CVPR_minet} & CVPR'20 & ResNet-50  & 82.03 & 0.847 & 0.092 & 8.55 & 85.88 & 0.881 & 0.06 & 6.03 & 76.61 & 0.798 & 0.104 & 10.33  \\
		ITSD$^\circ$ \cite{Zhou_2020_CVPR_ITSD} & CVPR'20 & ResNet-50 & 83.72 & 0.862 & 0.087 & 7.77 & 85.44 & 0.871 & 0.063 & 6.26 & 74.33 & 0.776 & 0.123 & 11.39 \\
		%    \hline
		%    PraNet \cite{fan2020pra_pranet} & MICCAI'20 & - & - & - & - \\
		%    \hline
		%    SINet \cite{Fan_2020_CVPR_sinet} & CVPR'20 & - & - & - & - \\
		\hline
		DSC$^\vartriangle$ \cite{Hu_2018_CVPR_dsc} & CVPR'18 & ResNet-50  & 83.56 & 0.855 & 0.090 & 7.97 & 86.37 & 0.882 & 0.058 & 5.65 & 74.79 & 0.773 & 0.119 & 11.14 \\
		BDRAR$^{\vartriangle\dag}$ \cite{Zhu_2018_ECCV_bdrar} & ECCV'18 & ResNet-50 & 80.01 & 0.847 & 0.098 & 9.87 & 85.00 & 0.870 & 0.061 & 6.04 & 75.32 & 0.802 & 0.101 & 11.13  \\
		\hline
		PraNet$^\S$ \cite{fan2020pra_pranet} & MICCAI'20 & ResNet-50 & 82.06 & 0.847 & 0.098 & 9.33 & 87.15 & 0.881 & 0.058 & 5.31 & 71.93 & 0.756 & 0.128 & 13.11  \\
		\hline
		MirrorNet*$^\dag$ \cite{yang2019mirrornet} & ICCV'19 & ResNeXt-101  & 85.07 & 0.866 & 0.083 & 7.67 & 88.30 & \textbf{\blue{0.907}} & 0.047 & 4.95 & \textbf{\blue{78.82}} & \textbf{\blue{0.820}} & 0.102 & 9.93 \\
		\hline
		TransLab$^\diamond$ \cite{xie2020segmenting} & ECCV'20 & ResNet-50  & 81.64 & 0.849 & 0.097 & 9.70 & 87.10 & 0.897 & 0.051 & 5.44 & 74.32 & 0.781 & 0.123 & 12.00  \\
		Trans2Seg$^\diamond$ \cite{xie2021segmenting} & IJCAI'21 & ResNet-50  & 84.41 & 0.872 & 0.078 & 7.36 & 74.98 & 0.767 & 0.124 & 10.73 & 77.98 & 0.817 & 0.095 & 9.65 \\
		\hline
		GDNet$^\star$ \cite{Mei_2020_CVPR} & CVPR'20 & ResNeXt-101  & \textbf{\green{87.63}} & \textbf{\green{0.898}} & \textbf{\green{0.063}} & \textbf{\green{5.62}} & 88.68 & \textbf{\blue{0.907}} & 0.046 & \textbf{\blue{4.72}} & 78.73 & 0.817 & 0.097 & \textbf{\green{9.32}}  \\
		GSD$^\star$ \cite{lin2021rich} & CVPR'21 & ResNeXt-101  & \textbf{\blue{87.53}} & \textbf{\blue{0.895}} & \textbf{\blue{0.066}} & 5.90 & \textbf{\green{89.67}} & \textbf{\red{0.917}} & \textbf{\red{0.042}} & \textbf{\green{4.52}} & \textbf{\green{78.86}} & 0.818 & 0.103 & 9.79 \\ 

		\hline
		\textbf{PGSNet}$^\star$  & Ours & ResNeXt-101  & \textbf{\red{87.81}} & \textbf{\red{0.901}} & \textbf{\red{0.062}} & \textbf{\red{5.56}} & \textbf{\red{89.79}} & \textbf{\red{0.917}} & \textbf{\red{0.042}} & \textbf{\red{4.39}} & \textbf{\red{80.06}} & \textbf{\red{0.836}} & \textbf{\red{0.089}} & \textbf{\red{9.08}} \\
		\hline
		\hline
		
	\end{tabular}
	\label{tab:comparison}
\end{table*}

\begin{table}[htbp]
	\centering
	\footnotesize
	\setlength{\tabcolsep}{4.0pt}
	\caption{\final{Comparison between our PGSNet and state-of-the-art glass segmentation methods on the GSD dataset \cite{lin2021rich}.}}
	\begin{tabular}{c|c|p{1.55cm}<{\centering}|p{0.6cm}<{\centering}p{0.6cm}<{\centering}p{0.6cm}<{\centering}p{0.6cm}<{\centering}}
	\hline
	\hline
		\multirow{2}{*}{Methods} & \multirow{2}{*}{Pub.'Year} & \multirow{2}{*}{Backbone} & \multicolumn{4}{c}{GSD \cite{lin2021rich}} \\
		\cline{4-7}
		& & & IoU$\uparrow$ & $F_\beta^w$$\uparrow$ & MAE$\downarrow$ & BER$\downarrow$ \\
		
		\hline
		% 		ITSD & CVPR'20  & ResNet-50 & 83.72 & 0.862 & 0.087 & 7.77 \\
		% 		MirrorNet & ICCV'19  & ResNeXt-101 & 85.07 & 0.866 & 0.083 & 7.67 \\
		TransLab \cite{xie2020segmenting} & ECCV'20 & ResNet-50 &  78.05 & 0.828 & 0.069 & 9.19  \\
		Trans2Seg \cite{xie2021segmenting} & IJCAI'21 & ResNet-50 & 79.65 & 0.839 & 0.069 & 8.21 \\
		GDNet \cite{Mei_2020_CVPR} & CVPR'20 & ResNeXt-101 & 82.51 & 0.857 & 0.058 & 6.41 \\
		GSD \cite{lin2021rich} & CVPR'21 & ResNeXt-101 &  83.64 & \textbf{0.903} & 0.055 & \textbf{6.12} \\
		PGSNet & Ours & ResNeXt-101 & \textbf{83.65} & 0.868 & \textbf{0.054} & 6.25 \\
	\hline
	\hline		
	\end{tabular}
	\label{tab:vgg_resnet_performance}
\end{table}

\subsubsection{Compared Methods}
\mhy{We validate the effectiveness of our method by comparing it with  \blue{26} methods selected from other related fields according to the following criteria: (\romannumeral1) classical architectures, (\romannumeral2) recently published, and (\romannumeral3) achieving state-of-the-art performance in the specific field. Specifically, we choose semantic segmentation methods ICNet \cite{Zhao_2018_ECCV_icnet}, PSPNet \cite{zhao2017pyramid}, DeepLabv3+ \cite{chen2017deeplab}, DenseASPP \cite{Yang_2018_CVPR_DenseASPP}, BiSeNet \cite{Yu_2018_ECCV_BiSeNet}, DANet \cite{Fu_2019_CVPR_DANet}, CCNet \cite{Huang_2019_ICCV_CCNet}, GFFNet \cite{li2020gated}, SFNet \cite{li2020semantic} and FaPN \cite{huang2021fapn}; salient object detection methods DSS \cite{HouPami19Dss}, PiCANet \cite{liu2018picanet}, RAS \cite{Chen_2018_ECCV_ras}, CPD \cite{Wu_2019_CVPR}, EGNet \cite{Zhao_2019_ICCV_egnet}, F3Net \cite{wei2019f3net}, MINet-R \cite{Pang_2020_CVPR_minet}, and ITSD \cite{Zhou_2020_CVPR_ITSD}; shadow detection methods DSC \cite{Hu_2018_CVPR_dsc} and BDRAR \cite{Zhu_2018_ECCV_bdrar}; medical image segmentation method PraNet \cite{fan2020pra_pranet}; mirror segmentation method MirrorNet \cite{yang2019mirrornet}; transparent object segmentation methods TransLab \cite{xie2020segmenting} and Trans2Seg \cite{xie2021segmenting}; and glass segmentation method GDNet \cite{Mei_2020_CVPR} and GSD \cite{lin2021rich}. For a fair comparison, we use either their publicly available codes or the implementations with recommended parameter settings. For each of the three datasets, all the models are retrained on the corresponding training set. Besides, all the prediction maps are evaluated with the same code.}

\begin{figure*}[t]
	\begin{center}
		\includegraphics[width = 1\linewidth]{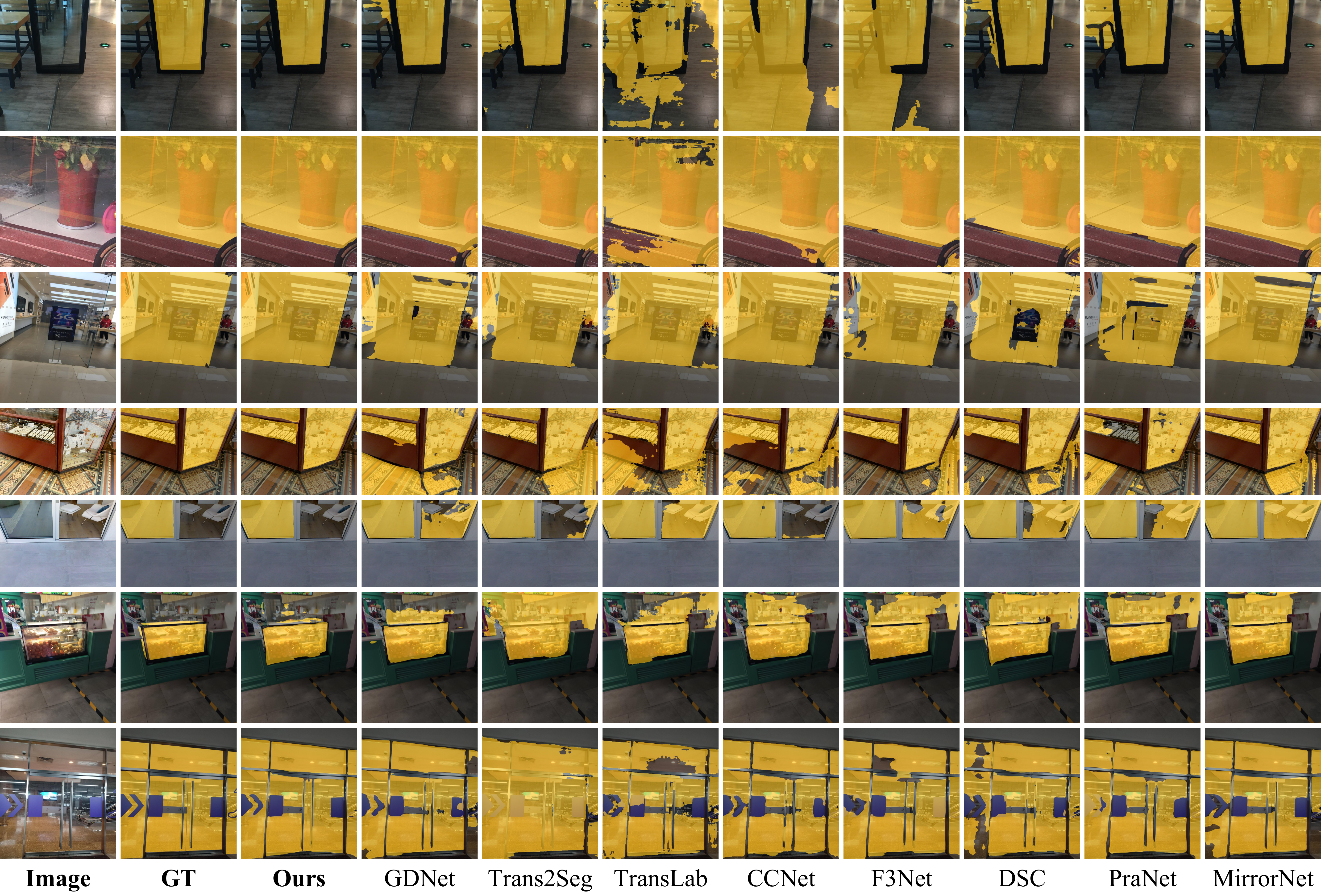}
	\end{center}
	\vspace{-12pt}
	\caption{Visual comparison of the proposed model with state-of-the-art methods. Obviously, our approach is capable of segmenting glass in various scenes more accurately.}
	\label{fig:visual}
	\vspace{-13pt}
\end{figure*}

\subsection{Comparison with the State-of-the-arts}
\mhy{Table \ref{tab:comparison} reports the quantitative results of our method against other  \newylt{26} state-of-the-art methods on two benchmark datasets and our newly constructed HSO dataset. We can see that our method outperforms all the other methods on all three challenging datasets under all four standard metrics. Notably, compared with the state-of-the-art glass segmentation method GDNet \cite{Mei_2020_CVPR}, our method improves $IoU$ and $F_\beta^w$ by $1.33\%$ and $1.90\%$ on the HSO dataset, respectively.} 
% \ylt{Our method is also faster than GDNet \cite{Mei_2020_CVPR}, \emph{i.e.}, 21 versus 21 FPS.}
\add{Noteworthy is that our HSO dataset leads to a \textasciitilde10\% IoU drop for all methods in Table \ref{tab:comparison}, reflecting the large room to achieve accurate glass segmentation and the necessity of the new HSO dataset for stimulating further research.}
\final{We also retrained our PGSNet on the GSD dataset \cite{lin2021rich} and presented the results in Table \ref{tab:vgg_resnet_performance}. It can be seen that our PGSNet achieves comparable performance against state-of-the-arts.}
\mhy{Besides, Figure \ref{fig:visual} qualitatively compares our PGSNet with three prior glass/transparency segmentation methods (\ie, GDNet \cite{Mei_2020_CVPR}, Trans2Seg \cite{xie2020segmenting}, and TransLab \cite{xie2020segmenting}) as well as the best approach from each of the six other categories (\ie, semantic segmentation method CCNet \cite{Huang_2019_ICCV_CCNet}, salient object detection method F3Net \cite{wei2019f3net}, shadow detection method DSC \cite{Hu_2018_CVPR_dsc}, medical image segmentation method PraNet \cite{fan2020pra_pranet}, and mirror segmentation method MirrorNet \cite{yang2019mirrornet}).
It can be seen that our method is capable of accurately segmenting small glass regions (\eg, 1\textit{st} row), large glass regions (\eg, 2-\textit{nd} and 3-\textit{rd} rows), and multiple glass regions (\eg, 4-\textit{th} row). This is mainly because that the discriminability enhancement (DE) module can explore different scales of contexts and purify the features to be the more discriminative representations, benefiting the more accurate positioning of glass regions. While the state-of-the-arts are typically confused by the background which shares similar appearance with the glass regions (\eg, 5-\textit{th} row) or the glass region that cluttered in the background (\eg, 6-\textit{th} row), our method can successfully infer the true glass region. This is mainly contributed by the proposed focus-and-exploration based fusion (FEBF) module which could help eliminate the ambiguous features and augment the details. Furthermore, benefited by the progressive features fusion from high-level to low-level, our method can effectively perceive the detailed information and thus has the ability to finely segment the glass regions with complex structures (\eg, the last row).}

\subsection{Ablation Study}
\add{Our work is motivated by the fact that existing glass segmentation methods ignore the importance of the effective fusion of different features/cues.  We design the DE module as a multi-/large-field processing paradigm based on the observation that a set of various sized population receptive fields helps to perceive the small spatial shifts and the contexts in a large receptive field could help to enhance the features semantic. The FEBF module is designed to alleviate both the introduction of ambiguous and loss of details in the fusion via highlighting the common and exploring the difference between level-different features, respectively.}

\begin{figure*}[tbp]
	\centering
	\includegraphics[width=1\linewidth]{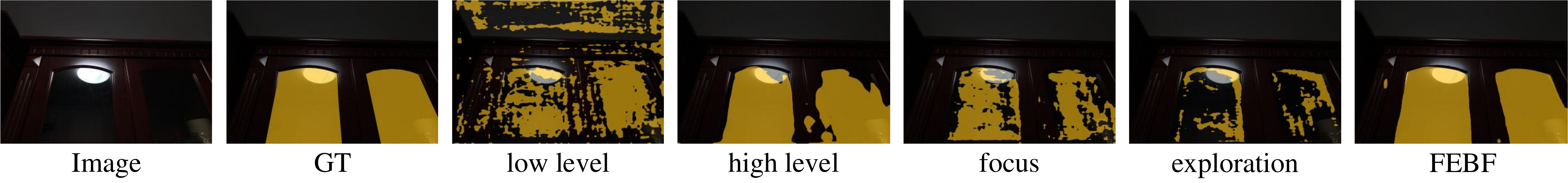}
% 	\caption{\add{Feature visualization in FEBF module. }}
    \vspace{-18pt}
    % \caption{\final{Visual results of low-level features, high-level features, features from focus branch, features from exploration branch and features from FEBF module for showing the effects on FEBF module.}}
    % Pls. do not change this.
    \caption{\final{Visual examples that reveal the rationality behind the FEBF module design. It visualizes both input low-level features (the 3\textit{rd} column) and high-level features (the 4\textit{th} column) as well as output features (the last column) of the FEBF module, together with the intermediate features passed through the focus branch (the 5\textit{th} column) and exploration branch (the 6\textit{th} column) in the FEBF module. Although the input low-level features are full of noise and the input high-level features are coarse, the focus branch can highlight the common and the exploration branch can explore the difference, making the final output features of the FEBF module more discriminative.}}
    % \caption{\final{Visual examples that reveal the rationality behind the FEBF module design. Although the input low-level features of the FEBF module (the 3\textit{rd} column) are full of noise and the input high-level features (the 4\textit{th} column) are coarse, the focus branch in the FEBF module can highlight the common (the 5\textit{th} column) and the exploration branch can explore the difference (the 6\textit{th} column), making the final output features of the FEBF module (the last column) more discriminative.}}
    \vspace{-10pt}
    \label{fig:visual_features}
\end{figure*}

\mhy{In this subsection, we conduct ablation studies to validate the effectiveness of two key components tailored for effective features fusion towards accurate glass segmentation, \ie, discriminability enhancement (DE) module and focus-and-exploration based fusion (FEBF) module, as well as explore the impact of different feature extractors, and report the experimental results in Table \ref{tab:ablation_study}, \ref{tab:comparison_with_general_attention}, \ref{tab:vgg_resnet_performancev2}, and Figure \ref{fig:visual_features}.}

\begin{table}[t]
	\centering
	\footnotesize
	\setlength{\tabcolsep}{4.0pt}
	\caption{\mhy{Quantitative ablation results that indicate that each component in PGSNet contributes to the overall performance. ``B'' denotes our network with our proposed feature fusion strategy (\ie, discriminability enhancement (DE) module and focus-and-exploration based fusion (FEBF) module) replaced by the concatenation fusion strategy. ``LFE'', ``LFF'', and ``CFP'' are the local feature extraction, local feature fusion, and contextual feature perception in DE module, respectively.}} 
	\begin{tabular}{cl|cccc}
		\hline
		\hline
		\multicolumn{2}{c|}{\multirow{2}{*}{Networks}} & \multicolumn{4}{c}{GDD \cite{Mei_2020_CVPR}} \\
		\cline{3-6}
		& & IoU$\uparrow$ & $F_\beta^w$$\uparrow$ & MAE$\downarrow$ & BER$\downarrow$ \\
		
		\hline
		$\mathit{A}$ & B (\ie, Concatenation Fusion) & 85.63 & 0.881 & 0.075 & 7.04 \\
		$\mathit{B}$ & B + DE \textit{w/} LFE only & 87.28 & 0.893 & 0.066 & 6.01 \\
		$\mathit{C}$ & B + DE \textit{w/} LFE\&LFF only & 87.57 & 0.896 & 0.065 & 5.94 \\
		$\mathit{D}$ & B + DE \textit{w/} LFE\&LFF\&CFP & 87.61 & 0.897 & 0.063 & 5.75 \\
		\hline
		$\mathit{E}$ & B \textit{w/} Addition Fusion & 86.02 & 0.885 & 0.071 & 6.84 \\
		$\mathit{F}$ & B \textit{w/} Multiplication Fusion & 85.88 & 0.885 & 0.073 & 6.90 \\
		$\mathit{G}$ & B + FEBF \textit{w/} Focus only & 86.22 & 0.888 & 0.070 & 6.72 \\
		% 		$\mathit{H}$ & B + FEBF \textit{w/} Exploration only & 85.86 & 0.886 & 0.072 & 6.89 \\
		$\mathit{H}$ & B + FEBF \textit{w/} Focus\&Exploration & 86.55 & 0.888 & 0.071 & 6.56 \\
		\hline
		$\mathit{I}$ & B + DE + FEBF (\ie, PGSNet) & \textbf{87.81} & {\textbf{0.901}} & \textbf{0.062} & \textbf{5.56} \\
		\hline
		\hline
	\end{tabular}
	\vspace{-2pt}
	\label{tab:ablation_study}
\end{table}

\subsubsection{The effectiveness of DE module}
\mhy{We first define and train a base model ``B'' (\ie, $\mathit{A}$ in Table \ref{tab:ablation_study}) which is based on PGSNet but the feature fusion strategy (\ie, DE module and FEBF module) is replaced by the concatenation fusion strategy. Starting from the base model, we gradually re-introduce the LFE, LFF, and CFP in the DE module ($\mathit{B}$-$\mathit{D}$). From the results, we observe that: (\romannumeral1) DE module can help boost the segmentation performance largely (\ie, $\mathit{D}$ is better than $\mathit{A}$) as the DE bridges the characteristic gap between level-different features and thus benefit the effective features fusion; (\romannumeral2) applying multi-field processing on the level-specific features before fusing them together is helpful (\ie, $\mathit{B}$ is better than $\mathit{A}$), indicating that multi-field processing is essential for making features more discriminative and more suitable for fusion; (\romannumeral3) channel recalibration \cite{hu2018squeeze} can help integrate local features (\ie, $\mathit{C}$ is better than $\mathit{B}$); and (\romannumeral4) perceiving contextual information can further improve the segmentation performance (\ie, $\mathit{D}$ is better than $\mathit{C}$).}
\add{Note that as a pre-step in features fusion, DE module does NOT aim to explore specific glass-related cues but to enhance the features discriminability for effective features fusion. It differs from the LCFI module \cite{Mei_2020_CVPR} in that it has the \textit{extra} local feature fusion (LFF) and contextual feature perception (CFP) which can effectively aggregate local features and enlarge the receptive field, respectively. The ablation study results (\textit{i.e.}, $\mathit{B}$-$\mathit{D}$ in Table \ref{tab:ablation_study}) validate the effectiveness of these two components.}
\add{Besides, the experimental results in Table \ref{tab:comparison_with_general_attention} further demonstrate the superiority of our proposed DE module over general attention modules (\textit{i.e.}, channel attention, spatial attention, and combination of the two).}

%\ylt{Compared with the general attention module in Table \ref{tab:comparison_with_general_attention}, we find that our proposed DE module can enhance the features discriminability and gather more useful information than general attention module, resulting in better performance.  }

%\ylt{Note that as a pre-step in features fusion, DE module does NOT aim to explore specific glass-related cues but to enhance the features discriminability for effective features fusion. It differs from the LCFI module\cite{Mei_2020_CVPR} in that it has the extra local feature fusion (LFF) and contextual feature perception (CFP) which can effectively aggregate local features and enlarge the receptive field, respectively.}

\begin{table}[tbp]
	\centering
% 	\footnotesize
	\setlength{\tabcolsep}{4.0pt}
	\caption{\add{Quantitative Ablation Results that demonstrate the superiority of our proposed DE module over general attention modules (\textit{i.e.}, channel attention, spatial attention, and combination of the two).}}
	\begin{tabular}{l|p{0.6cm}<{\centering}p{0.6cm}<{\centering}p{0.6cm}<{\centering}p{0.6cm}<{\centering}}
	    \hline
	    \hline
		\multirow{2}{*}{Networks} & \multicolumn{4}{c}{GDD \cite{Mei_2020_CVPR}} \\
		\cline{2-5}
		&  IoU$\uparrow$ & $F_\beta^w$$\uparrow$ & MAE$\downarrow$ & BER$\downarrow$ \\
		\hline
		PGSNet \textit{w/o} attention  & 86.55 & 0.888 & 0.071 & 6.56 \\
		PGSNet \textit{w/} channel attention & 86.60 & 0.892 & 0.066 & 6.51 \\
		PGSNet \textit{w/} spatial attention & 86.98 & 0.893 & 0.068 & 6.30 \\
		PGSNet \textit{w/} channel+spatial attention & 87.22 & 0.892 & 0.068 & 6.18\\
		\hline
		PGSNet \textit{w/} DE module  & \textbf{87.81} & \textbf{0.901} & \textbf{0.062} & \textbf{5.56} \\
	    \hline
	    \hline
	\end{tabular}
	\label{tab:comparison_with_general_attention}
\end{table}

\subsubsection{The effectiveness of FEBF module}
\mhy{Compared to the concatenation fusion strategy (\ie, $\mathit{A}$ in Table \ref{tab:ablation_study}), the strategies of addition ($\mathit{E}$) and multiplication ($\mathit{F}$) fusion  (\newylt{as shown in Figure \ref{fig:fusion} (a)}) perform similarly. \newylt{Due to characteristic gap between level-different features and redundant information in the fusion results, these simple fusion strategies would limit the accuracy of glass segmentation tasks.} By highlighting the common between level-different features in the fusion process (\ie, $\mathit{G}$, \newylt{also shown in Figure \ref{fig:fusion} (c)}), a better performance can be achieved. When further introduce the exploration of features difference (\ie, $\mathit{H}$, \newylt{also shown in Figure \ref{fig:fusion} (d)}), we get improved results. \newylt{Also shown in the fourth to seventh rows in
Figure \ref{fig:visual}, when the glass regions are in 
similar patterns/semantic information with surroundings or have strong reflections, our FEBF module is superior to the traditional attention or interaction mechanism.} This clearly demonstrates that our FEBF is more effective for features fusion towards accurate glass segmentation.}
\mhy{Finally, the combination of DE and FEBF module (\ie, $\mathit{I}$) enables our approach to possess the strong capability of accurately segmenting the glass.}

\add{We further visualize both input and output features of FEBF module in Figure \ref{fig:visual_features} to validate the rationality behind the module design. As we can see, the input low-level features are full of noise (the 3\textit{rd} column) and the input high-level features are coarse (inaccurate for the glass boundary region, the 4\textit{th} column). The focus branch in FEBF module can highlight the common (the 5\textit{th} column) and the exploration branch can explore the difference (the 6\textit{th} column) between low-level and high-level features, respectively. And the combination of features from both focus branch and exploration branch gets more pure and accurate glass features (the last column). In conclusion, our well-designed FEBF module can effectively alleviate the introduction of ambiguous features and the loss of details when aggregating different levels of features, and thus can facilitate the optimization of the whole network.}

\subsubsection{The impact of different feature extractors}
\add{We follow \cite{Mei_2020_CVPR} to use ResNeXt-101 \cite{xie2017aggregated} as the backbone to extract different levels of features. We further investigate the performance of more shallow/lightweight feature extractors and report the results in Table \ref{tab:vgg_resnet_performancev2}. We can observe that using more simple feature extractors (\textit{i.e.}, VGG-16 \cite{simonyan2014very} or ResNet-50 \cite{he2016deep_resnet}) will lead to the performance decline in some extent. However, our PGSNet still achieve better results than other SOTA methods under the same simple feature extractors, showing the superiority of our method over the others.}
%\ylt{Although we utilize the same ResNeXt-101 backbone \cite{xie2017aggregated} to extract backbone features. From Table \ref{tab:vgg_resnet_performance}, we observe that even using the weaker feature extractors \eg, VGG-16 \cite{simonyan2014very} or ResNet-50 \cite{he2016deep_resnet}, our proposed PDNet model can still reach a better performance than the newly conducted method TransLab\cite{xie2020segmenting} and Trans2Seg\cite{xie2021segmenting}, and outperform GDNet\cite{Mei_2020_CVPR} when both using ResNet-50 backbone for fairness comparisons.}

\begin{table}[tbp]
	\centering
	\footnotesize
	\setlength{\tabcolsep}{4.0pt}
	\caption{\add{Comparison between our PGSNet and other SOTA methods under different feature extractors.}}
	\begin{tabular}{c|c|p{1.55cm}<{\centering}|p{0.6cm}<{\centering}p{0.6cm}<{\centering}p{0.6cm}<{\centering}p{0.6cm}<{\centering}}
	\hline
	\hline
		\multirow{2}{*}{Methods} & \multirow{2}{*}{Pub.'Year} & \multirow{2}{*}{Backbone} & \multicolumn{4}{c}{GDD \cite{Mei_2020_CVPR}} \\
		\cline{4-7}
		& & & IoU$\uparrow$ & $F_\beta^w$$\uparrow$ & MAE$\downarrow$ & BER$\downarrow$ \\
		
		\hline
		% 		ITSD & CVPR'20  & ResNet-50 & 83.72 & 0.862 & 0.087 & 7.77 \\
		% 		MirrorNet & ICCV'19  & ResNeXt-101 & 85.07 & 0.866 & 0.083 & 7.67 \\
		TransLab \cite{xie2020segmenting} & ECCV'20 & ResNet-50 & 81.64 & 0.849 & 0.097 & 9.70 \\
		Trans2Seg \cite{xie2021segmenting} & IJCAI'21 & ResNet-50 & 84.41 & 0.872 & 0.097 & 7.36 \\
		GDNet \cite{Mei_2020_CVPR} & CVPR'20 & ResNet-50 & 85.23 & 0.877 & 0.074 & 6.65 \\
		\hline
		PGSNet & Ours & VGG-16 & 86.92 & 0.891 & 0.067 & 6.20 \\
		PGSNet & Ours & ResNet-50 & 86.88 & 0.890 & 0.067 & 6.00 \\
		PGSNet & Ours & ResNeXt-101 & \textbf{87.81} & \textbf{0.901} & \textbf{0.062} & \textbf{5.56} \\
	\hline
	\hline		
	\end{tabular}
	\label{tab:vgg_resnet_performancev2}
\end{table}

\subsection{Computational Cost}
% \add{Benefited by the proposed features fusion strategy, our PGSNet achieves better performance than the state-of-the-art glass segmentation method GDNet \cite{Mei_2020_CVPR} without increasing the computational cost. Table \ref{tab:flops_paramsv2} compares the superior computational efficiency of our PGSNet against GDNet \cite{Mei_2020_CVPR} in terms of FLOPs (in G) and model parameters (in M).}
% \newylt{Note that, we then resize the input images to 352$\times$352 during training and validation to retrain GDNet \cite{Mei_2020_CVPR}. As shown in Table \ref{tab:flops_paramsv2}, row 3,  the performance drops with the decrease of input size, however, our PGSNet is still more computationally efficient than GDNet \cite{Mei_2020_CVPR}.}
\final{We show the computational efficiency comparison between our PGSNet and state-of-the-art glass segmentation methods GDNet \cite{Mei_2020_CVPR} and GSD \cite{lin2021rich} in Table \ref{tab:flops_paramsv2}. Under each input resolution, our PGSNet needs less than half of the FLOPs compared to GDNet \cite{Mei_2020_CVPR} and slightly higher FLOPs than GSD \cite{lin2021rich}. Note that GDNet \cite{Mei_2020_CVPR}, GSD \cite{lin2021rich}, and our PGSNet achieve the best performance with the input scale of 416, 384, and 352, respectively, which means that our PGSNet is more efficient than the other two methods (\ie, 80.789 \textit{vs} 271.533/92.697) while has superior segmentation performance (please refer to the quantitative comparison results shown in  Table \ref{tab:comparison}).}

\begin{table}[htbp]
	\centering
	\footnotesize
	\setlength{\tabcolsep}{4.0pt}
	\caption{\final{Comparison of the computational efficiency between our PGSNet and state-of-the-art glass segmentation methods GDNet \cite{Mei_2020_CVPR} and GSD \cite{lin2021rich}. \textbf{*} indicates the resolution under which the best performance is achieved for each method.}}
	\begin{tabular}{p{1.8cm}<{\centering}|p{1.8cm}<{\centering}|p{1.8cm}<{\centering}|p{1.8cm}<{\centering}}

	\hline
	\hline
		\multirow{2}{*}{Methods}  & \multicolumn{3}{c}{FLOPs (G)}\\
		\cline{2-4}
		& Size: 352$\times$352 & Size: 384$\times$384 & Size: 416$\times$416 \\
		\hline
		GDNet \cite{Mei_2020_CVPR} & 191.411 & 231.365 & 271.533\textbf{*} \\
		\hline
        GSD \cite{lin2021rich} & 77.892 & 92.697\textbf{*} & 108.790 \\
        \hline
        PGSNet & 80.789\textbf{*} & 96.145 & 112.837  \\
	\hline
	\hline
	\end{tabular}
	\vspace{-10pt}
	\label{tab:flops_paramsv2}
\end{table}

% \subsection{Numbers of DE Branches}
% \newylt{We provide two different variants to show our contributions about the DE module. In the first experiment, DE module only contains one DE branch with the smallest 3$\times$3 convolution. In the second experiment, DE module contains two DE branches with the second smallest convolution (5$\times$5) and the smallest one (3$\times$3). We calculate the numbers of DE branches in each experiment and the corresponding results are in Table \ref{tab:receptive_field}. Our DE module contains larger receptive field with more DE branches, mines more diversified glass features, provides more satisfactory performance compared with other variants. }

\subsection{Influence of receptive field}
\final{Glass segmentation is different from appearance-formulation-based semantic segmentation and it heavily relies on the context. Towards accurate glass segmentation, existing state-of-the-art methods explore the context in diverse manners. For example, GDNet \cite{Mei_2020_CVPR} harvests large-field contexts via a well-designed LCFI module, TransLab \cite{xie2020segmenting} and Trans2Seg \cite{xie2021segmenting} leverages non-local operations to perceive long-range contexts, and GSD \cite{lin2021rich} aggregates rich contexts with the RCAM module. 
In our work, we also explore the context by implementing a DE module. Our DE module is based on the LCFI module \cite{Mei_2020_CVPR} but has two extra novel designs (\ie, local feature fusion LFF and contextual feature perception CFP). The LFF and CFP help further enlarge the receptive field and their effectiveness has been validated by our experiments (Table \ref{tab:ablation_study}).
As it is hard to calculate the exact receptive field of existing complicated networks, we instead further explore the influence of the receptive field on glass segmentation by conducting experiments that vary the number of DE branches to achieve different receptive fields.
From Table \ref{tab:receptive_field}, we can see that, despite possessing a larger receptive field, the PGSNet model with two DE branches performs similarly to the one with only one DE branch. We infer the reason behind this is that the performance of PGSNet is not tied to a single component but due to the interplay of all components combined. As such, PGSNet offers a holistic glass segmentation system that could cover the gap of small receptive field differences. When adopting four DE branches (\ie, the full PGSNet model), the receptive field becomes much larger than the former two variants which brings performance improvement to some extent. These experiments show that the receptive field indeed has an influence on glass segmentation and presents a the larger the better trend in a certain range.}

\begin{table}[tbp]
	\centering
	\footnotesize
	\setlength{\tabcolsep}{4.0pt}
% 	\renewcommand\arraystretch{0.8}
    % \vspace{-4pt}
	\caption{\newylt{Quantitative ablation results that reveal the influence of receptive fields.}}
	\begin{tabular}{c|cccc}
	\hline
	\hline
		\multirow{2}{*}{Networks}  & \multicolumn{4}{c}{GDD \cite{Mei_2020_CVPR}} \\
		\cline{2-5}
		&  IoU$\uparrow$ & $F_\beta^w$$\uparrow$ & MAE$\downarrow$ & BER$\downarrow$ \\
		
		\hline
		% 		ITSD & CVPR'20  & ResNet-50 & 83.72 & 0.862 & 0.087 & 7.77 \\
		% 		MirrorNet & ICCV'19  & ResNeXt-101 & 85.07 & 0.866 & 0.083 & 7.67 \\
		
		\hline
		PGSNet \textit{w/} one DE branch  & 86.85 & 0.895 & 0.067 & 6.28\\
		PGSNet \textit{w/} two DE branches  & 86.86 & 0.898 & 0.069 & 6.17 \\
		PGSNet \textit{w/} four DE branches  & \textbf{87.81} & \textbf{0.901} & \textbf{0.062} & \textbf{5.56} \\
	\hline
	\hline		
	\end{tabular}
	\vspace{-6pt}
	\label{tab:receptive_field}
\end{table}

% \vspace{-12pt}
\subsection{Loss function and training strategy}
\newylt{We use the same loss function as \cite{Mei_2020_CVPR} to optimize or network the same multi-scale supervision training strategy from \cite{wei2019f3net}. To verify the importance of the lesser-used IoU loss and the multi-scale training strategy, we compare our PGSNet with the ones without IoU loss for training and without multi-scale training strategy in Table \ref{tab:loss_strategy}. We investigate the contribution of multi-scale training strategy, we find that the multi-scale training strategy has little impact on the learning capabilty of the network, while only increase the IoU performance from 87.76 to 87.81 on GDD dataset \cite{Mei_2020_CVPR}. Compared with multi-scale training strategy, IoU loss seems playing a vital role for boosting the performance. That's a misunderstanding because the IoU loss only has ability to optimize the IoU terms in evaluation metrics for our PGSNet, when it comes to F-measure and MAE metrics, IoU loss term enhances the performances slightly. Notably, the IoU loss does not always provide performance boosting. Another ablation study on IoU loss when training Trans2Seg \cite{xie2021segmenting} shows that the IoU metrics decrease from 84.4 to 83.66 on GDD \cite{Mei_2020_CVPR} after increasing the auxiliary IoU loss.  }

\begin{table}[htbp]
	\centering
	\footnotesize
	\setlength{\tabcolsep}{4.0pt}
    \vspace{-6pt}
	\caption{\newylt{Comparison between our PGSNet and the one training without IoU loss or without multi-scale training strategy.}}
	\begin{tabular}{c|cccc}
	\hline
	\hline
		\multirow{2}{*}{Networks} & \multicolumn{4}{c}{GDD \cite{Mei_2020_CVPR}} \\
		\cline{2-5}
		& IoU$\uparrow$ & $F_\beta^w$$\uparrow$ & MAE$\downarrow$ & BER$\downarrow$ \\
		
		\hline
		% 		ITSD & CVPR'20  & ResNet-50 & 83.72 & 0.862 & 0.087 & 7.77 \\
		% 		MirrorNet & ICCV'19  & ResNeXt-101 & 85.07 & 0.866 & 0.083 & 7.67 \\
% 		TransLab \cite{xie2020segmenting}  & 81.64 & 0.849 & 0.097 & 9.70 \\
% 		TransLab \cite{xie2020segmenting} w/ IoU loss  & - & - & - & - \\
		Trans2Seg \cite{xie2021segmenting}  & 84.41 & 0.872 & 0.097 & 7.36 \\
        Trans2Seg \cite{xie2021segmenting} \textit{w/} IoU loss & 83.66 & 0.863 & 0.083 & 7.67 \\
		\hline
		PGSNet \textit{w/o} IoU loss & 86.97 & 0.898 & 0.066 & 6.04 \\
		PGSNet \textit{w/o} multi-scale training strategy  & 87.76 & 0.899 & 0.061 & 5.68 \\
		PGSNet  & \textbf{87.81} & \textbf{0.901} & \textbf{0.062} & \textbf{5.56} \\
	\hline
	\hline		
	\end{tabular}
	\vspace{-3pt}
	\label{tab:loss_strategy}
\end{table}

\subsection{\add{Application to other tasks}}

\add{With the help of the well-designed discriminability enhancement (DE) module and focus-and-exploration based fusion (FEBF) module, our PGSNet can effectively fuse different levels of features, and thus has the potential to handle other challenging vision task. In this subsection, we consider three tasks, \textit{i.e.}, transparency segmentation, mirror segmentation, and salient object detection, and conduct corresponding experiments to demonstrate the effectiveness and generalization capability of our proposed PGSNet.}

\subsubsection{Transparency Segmentation}
\add{Transparency segmentation aims to segment transparent regions in the scene. We retrain our PGSNet and conduct the performance evaluation on the Trans10K dataset \cite{xie2020segmenting} which contains both ``Stuff'' such as window, showcase, and glass guardrail, as well as ``Things'' such as eyeglass, cup, and bottle. We name the PGSNet retrained for transparency segmentation as PGSNet$^T$. We compare PGSNet$^T$ with three state-of-the-art transparency/glass segmentation methods, including GDNet \cite{Mei_2020_CVPR}, TransLab \cite{xie2020segmenting} and Trans2Seg \cite{xie2021segmenting}, and report the quantitative comparison results in Table \ref{tab:trans10k}. The superior results of our PGSNet$^T$ clearly demonstrate the effectiveness of our method for transparency segmentation. Besides, by comparing the overall segmentation results in Table \ref{tab:trans10k} and \ref{tab:comparison}, we can observe that the overall performance for both ``Stuff'' and ``Things'' segmentation is better than that of for ``Stuff'' only segmentation (\textit{e.g.}, 89.35 versus 87.10 in terms of IoU for TransLab \cite{xie2020segmenting} on both ``Stuff'' and ``Things'' and ``Stuff'' only segmentation). This shows that the ``Stuff'' is more challenging than ``Things'' to be segmented. We think the reason is that the ``Stuff'' typically has a larger size and contains more diversity in patterns.}

\vspace{-8pt}
\begin{table}[htbp]
	\centering
	\setlength{\tabcolsep}{4.0pt}
	\caption{\add{The quantitative evaluation on the transparency segmentation task.}}
	\begin{tabular}{c|p{1.5cm}<{\centering}|p{0.9cm}<{\centering}p{0.9cm}<{\centering}p{0.9cm}<{\centering}p{0.9cm}<{\centering}}
		\hline
		\hline
% 		Methods & Pub.'Year & IoU$\uparrow$ & $F_\beta^w$$\uparrow$ & MAE$\downarrow$ & BER$\downarrow$ \\
		\multirow{2}{*}{Methods} & \multirow{2}{*}{Pub.'Year}  & \multicolumn{4}{c}{Trans10K \cite{xie2020segmenting}} \\
		\cline{3-6}
		& & IoU$\uparrow$ & $F_\beta^w$$\uparrow$ & MAE$\downarrow$ & BER$\downarrow$ \\
		\hline
		GDNet \cite{Mei_2020_CVPR} & CVPR'20  & 91.72 & 0.933  & 0.027 & 3.04  \\
		TransLab\cite{xie2020segmenting} & ECCV'20 & 89.35 & 0.921 & 0.032 & 4.37\\
		Trans2Seg\cite{xie2021segmenting} & IJCAI'21 & 91.14 & 0.932  & 0.027  & 3.33 \\
		\hline
		PGSNet$^T$ & Ours & \textbf{92.60} & \textbf{0.940} & \textbf{0.025} & \textbf{2.81}  \\
		\hline
		\hline
	\end{tabular}
	\label{tab:trans10k}
\end{table}

%\ylt{We also use the full dataset not only the ``stuff'' category in Trans10K\cite{xie2020segmenting} to train and test our newly proposed method. We regard the ``stuff'' and the ``things'' labels as the glass masks to segment  both medium/large glass surfaces and small glass objects.  We compare our method with three state-of-the-art glass/transparent object segmentation methods, including GDNet\cite{Mei_2020_CVPR}, TransLab\cite{xie2020segmenting} and Trans2Seg\cite{xie2021segmenting}. The quantitative performance is reported in Table \ref{tab:trans10k}. Our method performs the best on the fully-used Trans10K dataset, verifies that our method can not only detect glass surfaces well but also segment the transparent objects accurately, which demonstrates the generalizability and robustness of our method for glass segmentation tasks. }

\subsubsection{Mirror Segmentation}
\add{Mirror segmentation aims to segment regions that belong to the mirror. We retrain our PGSNet and conduct the performance evaluation on the MSD dataset \cite{yang2019mirrornet} which is the first large-scale mirror segmentation dataset. We name the PGSNet retrained for mirror segmentation as PGSNet$^M$. We compare PGSNet$^M$ with the state-of-the-art mirror segmentation method MirrorNet \cite{yang2019mirrornet} and report the quantitative comparison results in Table \ref{tab:performance_mirror}. We can see that PGSNet$^M$ performs favorably against MirrorNet \cite{yang2019mirrornet}. Note that our PGSNet$^M$ is an end-to-end process that does not need any post-processing, unlike MirrorNet \cite{yang2019mirrornet} which require post-processing by a computationally costly fully connected conditional random field (CRF) \cite{krahenbuhl2011efficient}.}

%\ylt{In the area of mirror segmentation, we follow MirrorNet\cite{yang2019mirrornet}, a recent work, to train and test our method on MSD dataset \cite{yang2019mirrornet}. MSD dataset contains 3,063 images in the training set and 955 images in the testing set. Using the same configuration as MirrorNet\cite{yang2019mirrornet}, as shown in Table \ref{tab:performance_mirror}, our method achieves comparable performance with MirrorNet\cite{yang2019mirrornet}, which demonstrates our method has potential to be adopted in the mirror segmentation community. }

\begin{table}[htbp]
	\centering
	\footnotesize
	\setlength{\tabcolsep}{4.0pt}
	\caption{\add{The quantitative evaluation on the mirror segmentation task.}}
	\begin{tabular}{c|c|p{0.8cm}<{\centering}|p{0.85cm}<{\centering}|p{0.85cm}<{\centering}|p{0.85cm}<{\centering}}
		\hline
		\hline
		
		\multirow{2}{*}{Methods} & \multirow{2}{*}{Backbone} & \multicolumn{4}{c}{MSD \cite{yang2019mirrornet}} \\
		\cline{3-6}
		& &IoU$\uparrow$ & $F_\beta^w$$\uparrow$ & MAE$\downarrow$ & BER$\downarrow$ \\
		
		\hline
		MirrorNet \cite{yang2019mirrornet} & ResNeXt-101 & 78.88 & 0.841 &  0.066 & \textbf{6.43} \\
		\hline
		PGSNet$^M$ & ResNeXt-101 & \textbf{80.77} & \textbf{0.879} & \textbf{0.052} & 6.77\\
		\hline
		\hline
	\end{tabular}
	\label{tab:performance_mirror}
\end{table}

\subsubsection{Salient Object Detection}
\add{Salient object detection (SOD) is a fundamental yet challenging vision task which aims to highlight and segment the most visually distinctive objects in an input image. We retrain our PGSNet on DUTS-TR \cite{Wang_2017_CVPR_duts} and conduct the performance evaluation on two challenging datasets, \textit{i.e.}, DUT-OMRON \cite{yang2013saliency} and DUTS-TE \cite{Wang_2017_CVPR_duts}. We name the PGSNet retrained for salient object detection as PGSNet$^S$. We compare PGSNet$^S$ with three state-of-the-art SOD methods, including F3Net \cite{wei2019f3net}, MINet \cite{Pang_2020_CVPR_minet}, and ITSD \cite{Zhou_2020_CVPR_ITSD}, and report the quantitative comparison results in Table \ref{tab:performance_sod}. For a fair comparison, PGSNet$^S$ adopts the same feature extractor (\textit{i.e.}, ResNet-50 \cite{he2016deep_resnet}) as three compared SOD methods. From the results, we can see that our PGSNet$^S$ achieves comparable or even better performance than the compared methods in terms of four standard evaluation metrics, \textit{i.e.}, structure-measure ($S_\alpha$) \cite{fan2017structure}, adaptive E-measure ($E_\phi^a$) \cite{fan2018enhanced}, weighted F-measure ($F_\beta^w$), and mean absolute error ($M$). This can be strong evidence to demonstrate the generalization capability of our method.}

%\ylt{In this subsection, we apply our method to other binary segmentation tasks, \eg, salient object detection, mirror segmentation, to verify the effectiveness of our method. }
%

%\ylt{In the area of salient object detection,  we evaluate our method on two benchmark datasets:  DUT-OMRON \cite{yang2013saliency} and DUTS \cite{Wang_2017_CVPR_duts}.  DUT-OMRON \cite{yang2013saliency} has 5,168 images with one or two objects in each image. Most of the foreground objects are structurally complex. DUTS \cite{Wang_2017_CVPR_duts} is currently the largest salient object detection dataset, which are divided into 10,553 training images (DUTS-TR) and 5,019 testing images (DUTS-TE). 
%We use DUTS-TR as the training set and others as testing sets. For the sake of fareness, we retrain our method with ResNet-50\cite{he2016deep_resnet} backbone with other methods also use this backbone to extract features. }
%
%\ylt{In Table \ref{tab:performance_sod}, we observe that: in comparison with two state-of-the-art RGB salient object detection methods, F3Net\cite{wei2019f3net} and ITSD\cite{Zhou_2020_CVPR_ITSD}, our method achieves a comparable or even the best performance in terms of the evaluation metrics on structure-measure ($S_\alpha$) \cite{fan2017structure}, adaptive E-measure ($E_\phi^a$) \cite{fan2018enhanced}, weighted F-measure ($F_\beta^w$), and mean absolute error ($M$).}
%
%\ylt{It can be verified that our method can transfer to the salient object detection field and achieve sound performances.}

\begin{table}[tbp]
	\centering
	\footnotesize
	\setlength{\tabcolsep}{2.4pt}
	\caption{\add{The quantitative evaluation on the salient object detection task.}}
	\begin{tabular}{p{1.4cm}<{\centering}|p{1.3cm}<{\centering}|p{0.55cm}<{\centering}|p{0.55cm}<{\centering}|p{0.55cm}<{\centering}|p{0.55cm}<{\centering}|p{0.55cm}<{\centering}|p{0.55cm}<{\centering}|p{0.55cm}<{\centering}|p{0.55cm}<{\centering}}
		\hline
		\hline
		{\multirow{2}{*}{Methods}} & {\multirow{2}{*}{Backbone}} & \multicolumn{4}{c|}{DUT-OMRON\cite{yang2013saliency}} & \multicolumn{4}{c}{DUTS-TE\cite{Wang_2017_CVPR_duts}} \\
		\cline{3-10}
		& & $S_\alpha$$\uparrow$ & $E_\phi^{a}$$\uparrow$ & $F_\beta^w$$\uparrow$ & M$\downarrow$ & $S_\alpha$$\uparrow$ & $E_\phi^{a}$$\uparrow$ & $F_\beta^w$$\uparrow$ & M$\downarrow$ \\
		\hline
		F3Net \cite{wei2019f3net} & ResNet-50 & .838 & .870 & .747 & .053 & .888 & .902 & .835 & .035\\
 		MINet \cite{Pang_2020_CVPR_minet} & ResNet-50 & .834 & .866 & .737 & .056 & .885 & .899 & .824 & .037 \\ 
		ITSD \cite{Zhou_2020_CVPR_ITSD} & ResNet-50 & \textbf{.840} & .863 & .750 & .061 & .885 & .895 & .824 & .041\\
		\hline
		PGSNet$^S$ & ResNet-50 & .839	& \textbf{.878} & \textbf{.758} & \textbf{.048} & \textbf{.889}	& \textbf{.920}	& \textbf{.849}	& \textbf{.034} \\
		\hline
		\hline
	\end{tabular}
	\label{tab:performance_sod}
\end{table}

% \vspace{-9.5pt}
\subsection{Discussion}
\subsubsection{\final{Negative examples}} 
\begin{figure}[hbp]
	\begin{center}
		\includegraphics[width=1\linewidth]{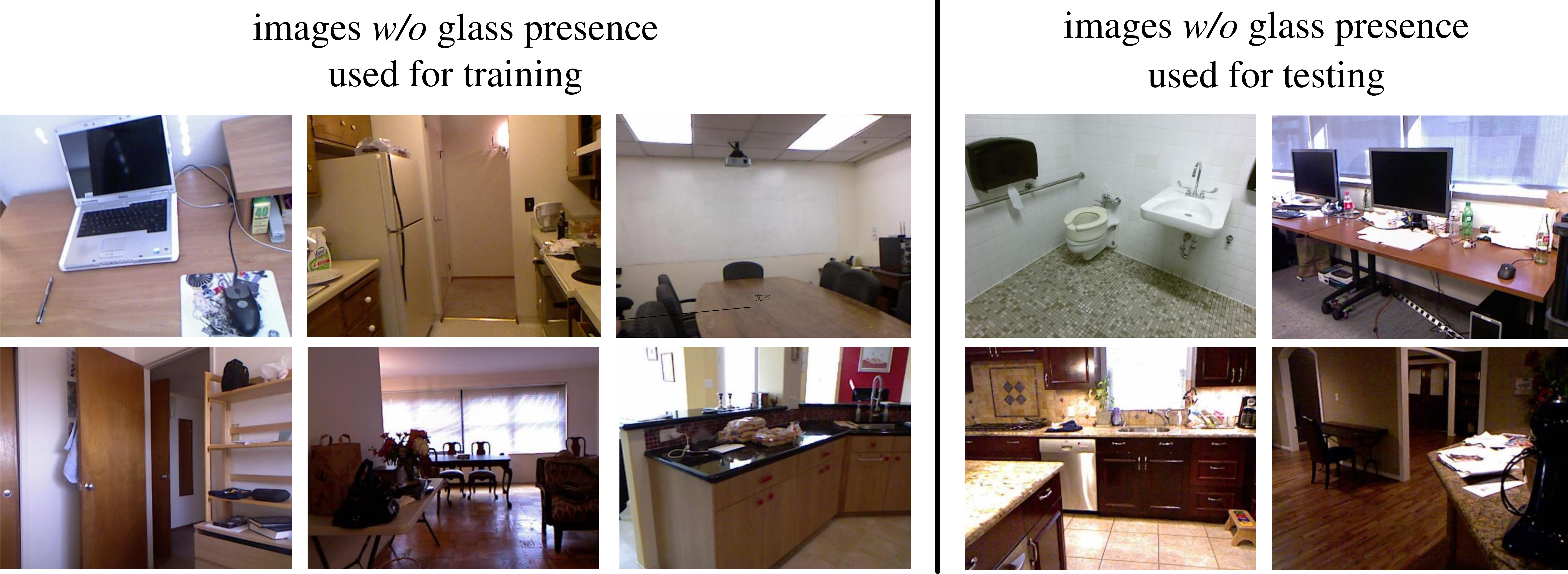}
	\end{center}
	\vspace{-10pt}
	\caption{\final{Visual examples of our curated negative set (\ie, images without glass presence).}}
	\vspace{-12pt}
	\label{fig:negative_set}
\end{figure}

\begin{figure}[tbp]
	\begin{center}
		\includegraphics[width=1\linewidth]{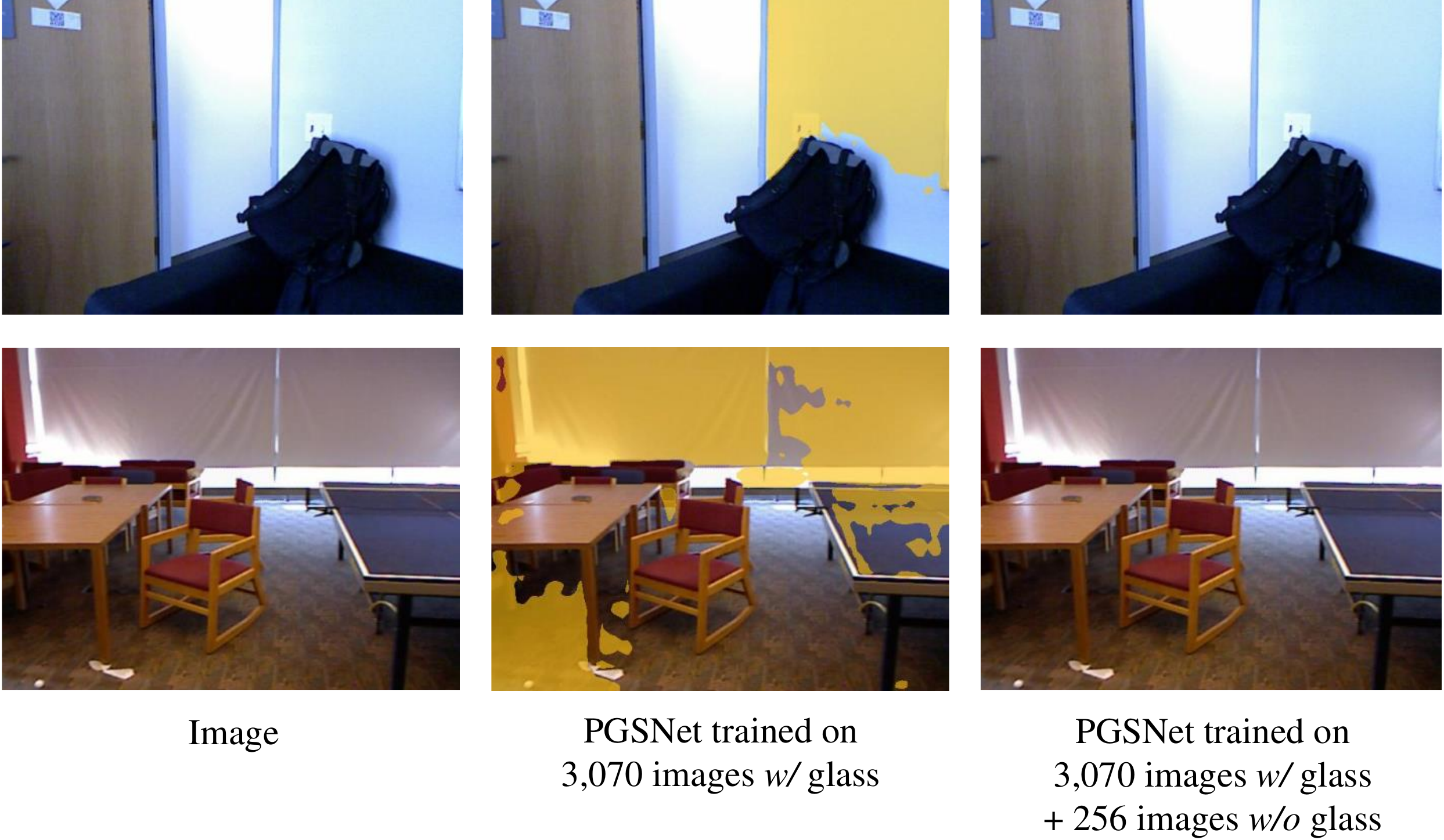}
	\end{center}
	\vspace{-12pt}
	\caption{\final{Our PGSNet trained with glass images only tends to over-segment the regions where perceptible boundary cues and highlights/reflections exist in the image without glass presence (\eg, the second column). Introducing the negative examples (\ie, images without glass presence) into the training process significantly reduces the false positive predictions (\eg, the third column).}}
	\label{fig:negative}
\end{figure}

\final{We first curate 367 images without glass presence from the widely used publicly available dataset (\ie, SUNRGBD\cite{song2015sun}) and show some examples in Figure \ref{fig:negative_set}. Then we conduct experiments based on images with and without glass presence and report the corresponding results in Table \ref{tab:performance_negative} and Figure \ref{fig:negative}. We can see that introducing the images without glass into the training process will decrease the testing performance on images with glass to some extent but can significantly reduce the false positive predictions on images without glass presence. Note that only the MAE results are presented for images without glass as the other three metrics are not applicable for the all-zero ground truth map.}

% All the exemplars in our HSO dataset contains at least one glass object, just like other existing glass dataset. To match the practical applications, we further include non-glass images. We random select 367 non-glass images from SUNRGBD\cite{song2015sun} dataset, dubbed HSO-N, and enlarge our origin HSO dataset to a new glass segmentation dataset  which both contains glass and non-glass images.  We retrain our PGSNet on the new constructed dataset and the results are listed as follows in Table \ref{tab:performance_negative}.

% As HSO-N only contains images without glass regions, the corresponding labels also don't contain glass labels,  so that we only compare MAE metrics for non-glass images. We can see that training with negative set will lead to a slight decrease on positive set (HSO), however, when it comes to negative set, the MAE metrics decrease from 0.304 to 0.000, which shows that introducing the negative examples into the training process significantly reduces the false positive prediction. In Figure \ref{fig:negative}, we can also find that training with negative examples will 
% eliminate the false segmentation of non-glass areas and strengthen the robustness of our PGSNet.
\vspace{-6pt}

% \vspace{1pt}
\begin{table}[htbp]
	\centering
	\footnotesize
	\setlength{\tabcolsep}{2.4pt}
	\caption{\final{The quantitative evaluation on images \textit{w/} and \textit{w/o} glass.}}
    \begin{tabular}{c|c|c|c|c|c}
		\hline
		\hline
		\multirow{2}{*}{Models} & \multicolumn{4}{c|}{\tabincell{c}{testing\\1782 images \textit{w/} glass}} & \tabincell{c}{testing\\111 images \textit{w/o} glass} \\
		\cline{2-6}
		& IoU$\uparrow$ & $F_\beta^w$$\uparrow$ & MAE$\downarrow$ & BER$\downarrow$ & MAE$\downarrow$ \\
		\hline
		\tabincell{c}{PGSNet trained on\\3070 images \textit{w/} glass} & \textbf{80.06} & \textbf{0.836} & 0.089 & \textbf{9.08} & 0.304	  \\
		\hline
		\tabincell{c}{PGSNet trained on\\3070 images \textit{w/} glass +\\256 images \textit{w/o} glass} & 79.84 & 0.831 & \textbf{0.086} & 9.29 & \textbf{0.000}	  \\
		\hline
		\hline
	\end{tabular}
	\label{tab:performance_negative}
\end{table}

% \begin{table}[htbp]
% 	\centering
% 	\footnotesize
% 	\setlength{\tabcolsep}{2.4pt}
% 	\caption{\final{The quantitative evaluation with/without a negative set on the HSO dataset.}}
% 	\begin{tabular}{p{3.4cm}<{\centering}|p{0.66cm}<{\centering}|p{0.66cm}<{\centering}|p{0.66cm}<{\centering}|p{0.66cm}<{\centering}|p{1.8cm}<{\centering}}
% 		\hline
% 		\hline
% 		{\multirow{3}{*}{PGSNet}} & \multicolumn{4}{c|}{HSO } & negative set \\
% 		&  \multicolumn{4}{c|}{1,782 test images} &
% 		 111 test images  \\
% 		\cline{2-6}
% 		& IoU$\uparrow$ & $F_\beta^w$$\uparrow$ & MAE$\downarrow$ & BER$\downarrow$ & MAE$\downarrow$ \\

% 		\hline
% 		HSO (3,070 train images) & 80.06 & 0.836 & 0.089 & 9.08 &  0.304	  \\
% 		\hline
% 		HSO + negative set & \multirow{2}{*}{79.84} & \multirow{2}{*}{0.831} & \multirow{2}{*}{0.086} &
% 		\multirow{2}{*}{9.29} & \multirow{2}{*}{0.000}	\\
% 		(3,326 train images) & & & &  \\
% 		\hline
% 		\hline
% 	\end{tabular}
% 	\label{tab:performance_negative}
% \end{table}

\subsubsection{\final{Limitation}}
\begin{figure}[tbp]
	\begin{center}
		\includegraphics[width=1\linewidth]{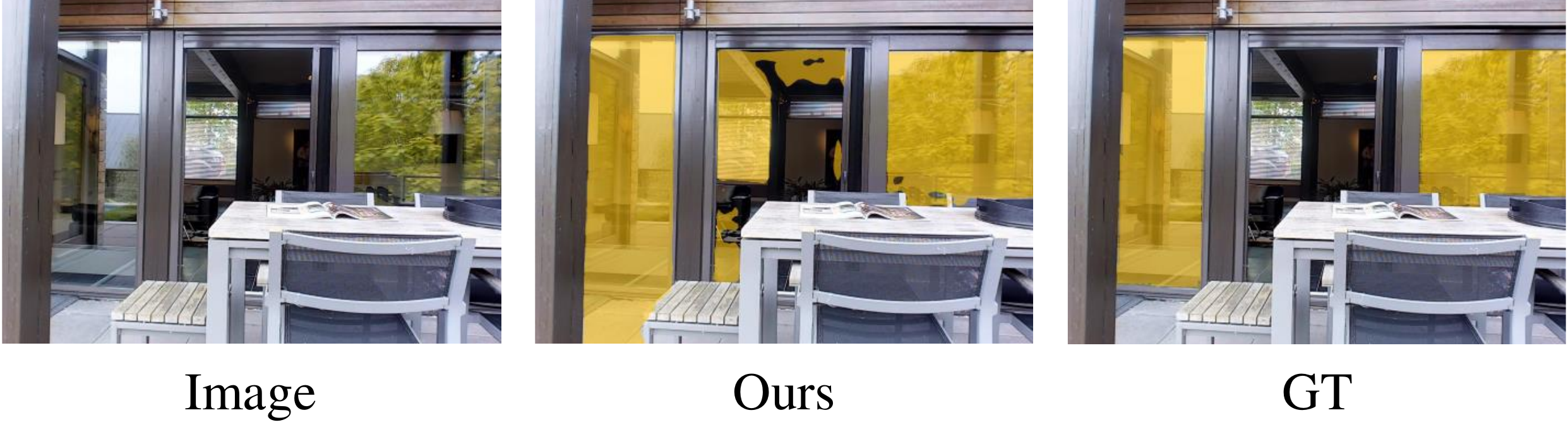}
	\end{center}
	\vspace{-12pt}
	\caption{\final{Failure case. Our method would fail to segment glass in some complex scenes where both high-level and low-level features can not contribute strong cues for distinguishing glass and non-glass regions. In this case, floors and window screens in the house share similar high-level contextual relations and low-level boundary/reflection cues to the true glass regions, which makes our our level-different features fusion based method naturally generate sub-optimal glass segmentation results.}}
	\label{fig:failure}
\end{figure}

\final{Note that the main focus of our work is about accuracy and our PGSNet is not a very efficient model. Besides, our method would fail in challenging scenes where both high-level and low-level features can not contribute strong cues for distinguishing glass and non-glass regions (\eg, as shown in Figure \ref{fig:failure}, floors and window screens in the house share similar high-level contextual relations and low-level boundary/reflection cues to the true glass regions, which makes our method output false positive predictions). A promising future research direction is to explore multi-view cues for robust glass segmentation in such complex scenes.}

\subsubsection{\final{Future work}} 
\final{Our future work is three-fold. First, our PGSNet still has a burden on computational efficiency, and it will be a great improvement by simplifying the PGSNet. Second, as the reflection information, a crucial cue for glass segmentation may be more obvious in the video, we are interested in glass segmentation in video fields. Finally, existing glass segmentation data is only based on static images. However, glass segmentation in other modalities (\eg, event data or depth information) raises new challenges under specific scenes.
}

% \vspace{-7pt}
\section{Conclusion}
\mhy{In this paper, we strive to embrace challenges in level-different features fusion towards accurate glass segmentation. We develop a novel feature fusion strategy for richly excavating useful information via both focus and exploration between features. By adopting this strategy in our progressive glass segmentation network (PGSNet), we show that our approach achieves state-of-the-art performance on two benchmark datasets as well as our home-scene-oriented (HSO) glass segmentation dataset. In the future, we plan to explore the potential of our method for other applications such as material segmentation and further enhance its capability for segmenting glass in videos.}
% \newylt{Another future work is about our HSO dataset. Our dataset doesn't contain a negative set in the evaluation data, where the images don't contain any glass. Although many datasets constructed for binary segmentation tasks, for example, DUTS-TE \cite{Wang_2017_CVPR_duts} for saliency detection, SBU \cite{vicente2016large, hou2019large} for shadow detection and MSD \cite{yang2019mirrornet} for mirror segmentation do not contain such negative sets and are used as benchmarks, a large dataset with negative samples will have more potentials on improving the robustness and accuracy for the practical applications. We will collect and enrich the negative samples in the future to strengthen our HSO dataset.}

\section*{Acknowledgment}
This work was supported in part by the National Natural Science Foundation of China under Grant  61972067/U21A20491/U1908214, National Key Research and Development Program of China (2021ZD0112400), and the Innovation Technology Funding of Dalian (2020JJ26GX036).

\ifCLASSOPTIONcaptionsoff
  \newpage
\fi

\bibliographystyle{IEEEtran}

\bibliography{paper}
%

% \clearpage
%
\begin{IEEEbiography}[{\includegraphics[width=1in,height=1.25in,clip,keepaspectratio]{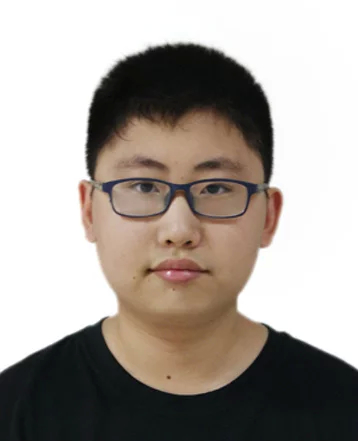}}]{Letian~Yu}
is a Ph.D. student in the School of Computer Science at Dalian University of Technology. He received his B.S. degree in Computer Science and Technology in 2019. His research interests are computer vision and image processing.
\end{IEEEbiography}

\begin{IEEEbiography}[{\includegraphics[width=1in,height=1.25in,clip,keepaspectratio]{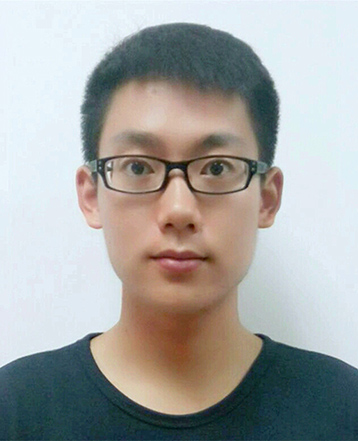}}]{Haiyang~Mei}
received the B.Eng. degree in the Automation from Dalian University of Technology, Dalian China, in 2017. He is currently working toward the Ph.D. degree with the School of Computer Science and Technology, Dalian University of Technology, Dalian, China. His research interests include image processing and computer vision.
\end{IEEEbiography}

\begin{IEEEbiography}[{\includegraphics[width=1in,height=1.25in,clip,keepaspectratio]{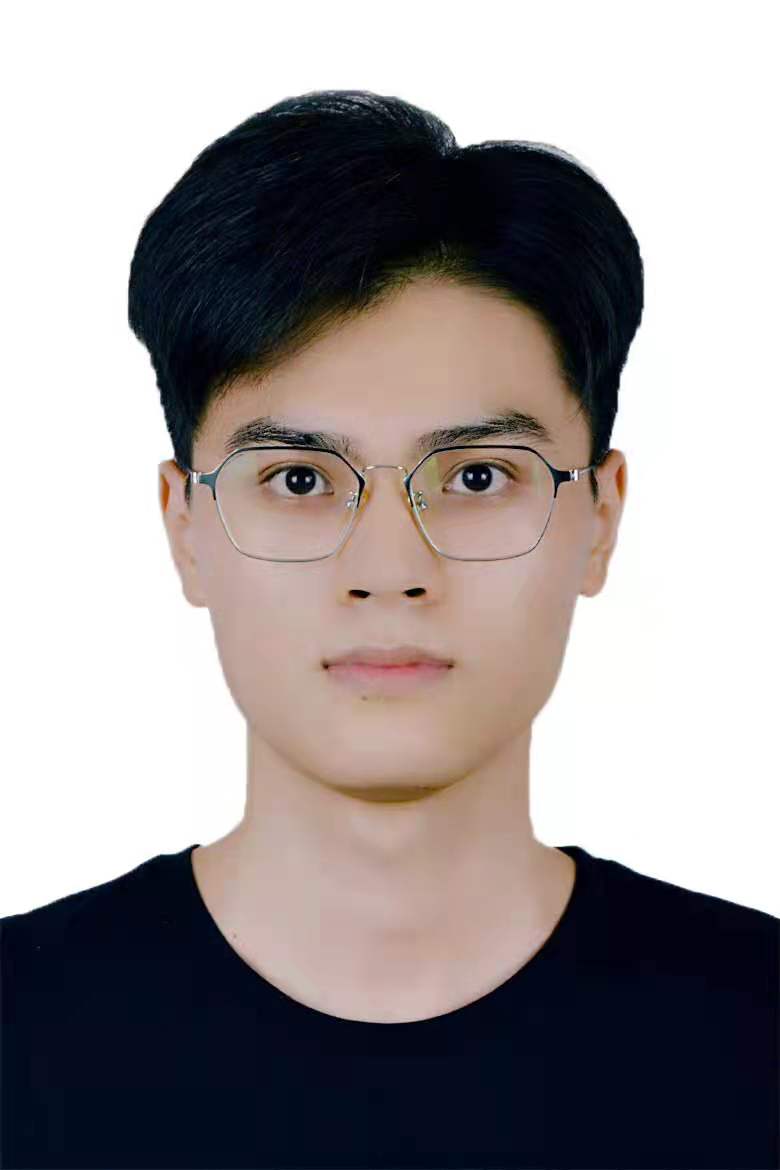}}]{Wen~Dong}
received his bachelor's degree in software engineering from Northwest A \& F University in 2019 and now is studying for a master's degree in the School of Computer Science and Technology at Dalian University of Technology. His research interests are computer vision and image processing.
\end{IEEEbiography}

\begin{IEEEbiography}[{\includegraphics[width=1in,height=1.25in,clip,keepaspectratio]{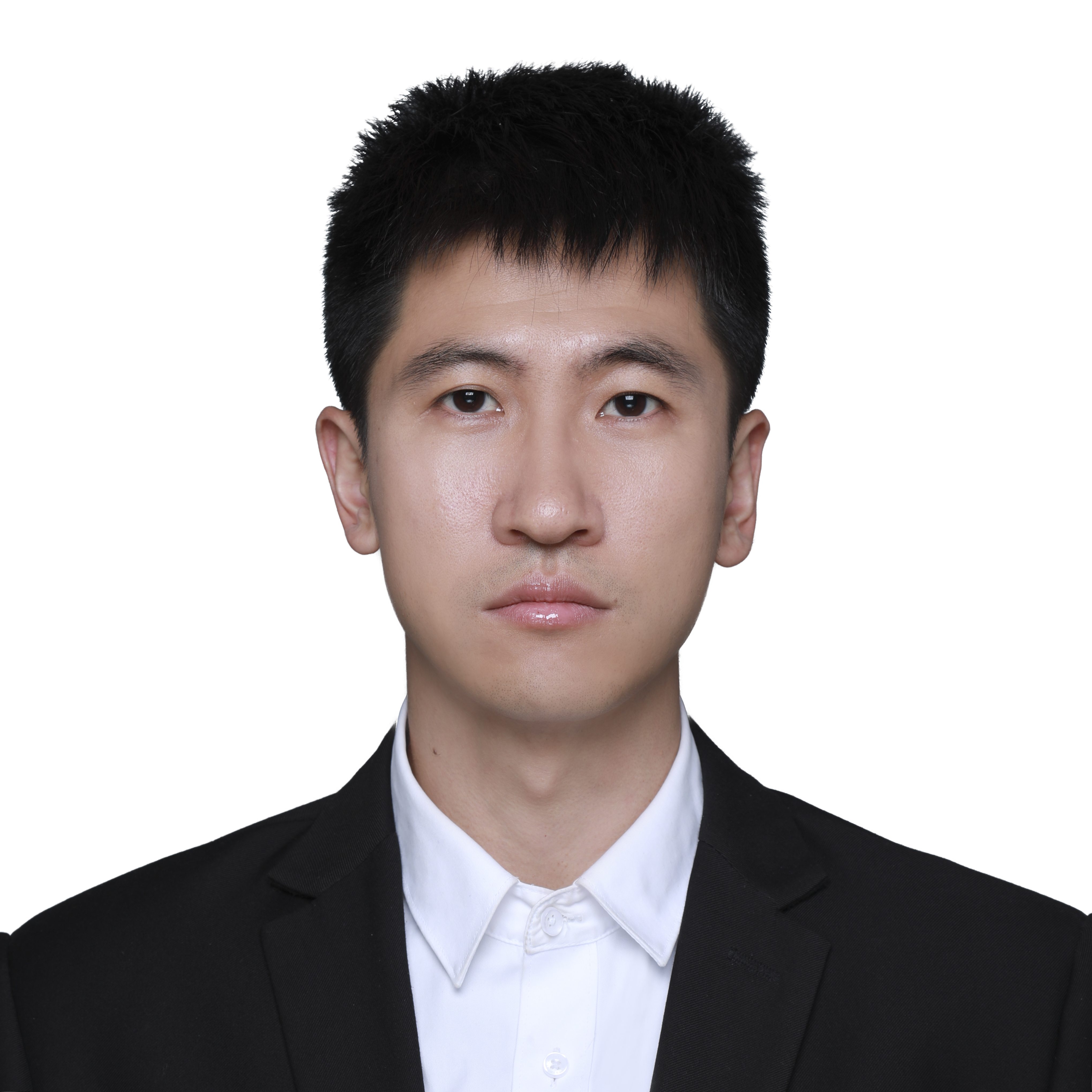}}]{Ziqi~Wei}
received the BS degree in computing science and technology from Renmin University of China, in June 2009 and the Ph.D. degree from University of Alberta, in Nov 2018. He is doing his post doc research in Tsinghua University. His research areas include wireless sensor network, computing theory, heuristic algorithms, health ageing and big data techniques in health care.
\end{IEEEbiography}

\vspace{-3cm}
\begin{IEEEbiography}[{\includegraphics[width=1in,height=1.25in,clip,keepaspectratio]{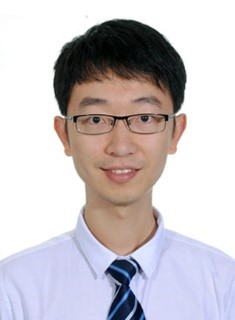}}]{Li~Zhu}
received the B.E. degree in automation from Dalian University of Technology, Dalian, China, in 2009, and the Ph.D. degree in control science and engineering from Zhejiang University, Hangzhou, China, in 2014.
He is currently an Assistant Professor with the School of Control Science and Engineering, Dalian University of Technology, Dalian, China. His current research interests include data mining and analytics, process monitoring and diagnosis, machine intelligence, and knowledge automation.
\end{IEEEbiography}

\vspace{-3cm}
\begin{IEEEbiography}[{\includegraphics[width=1in,height=1.25in,clip,keepaspectratio]{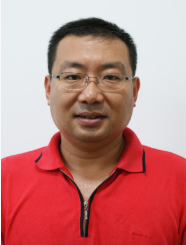}}]{Yuxin Wang}
received the MS degree in Computer Application in 2000 and the PhD degree in Computer Science in 2012 from Dalian University of Technology, P.R. China. He is currently an Associate Professor in the College of Computer Science at Dalian University of Technology. His research interests include distributed computing, big data analysis, deep learning and computer vision.
\end{IEEEbiography}

% \begin{IEEEbiography}[{\includegraphics[width=1in,height=1.25in,clip,keepaspectratio]{biography/baocaiyin}}]{Baocai~Yin}
% is a Professor of Computer Science at Dalian University of Technology and the Dean of the Faculty of Electronic Information and Electrical Engineering. His research concentrates on digital multimedia and computer vision. He received his B.S. degree and Ph.D. degree in Computer Science, each from Dalian University of Technology.
% \end{IEEEbiography}

\vspace{-3cm}
\begin{IEEEbiography}[{\includegraphics[width=1in,height=1.25in,clip,keepaspectratio]{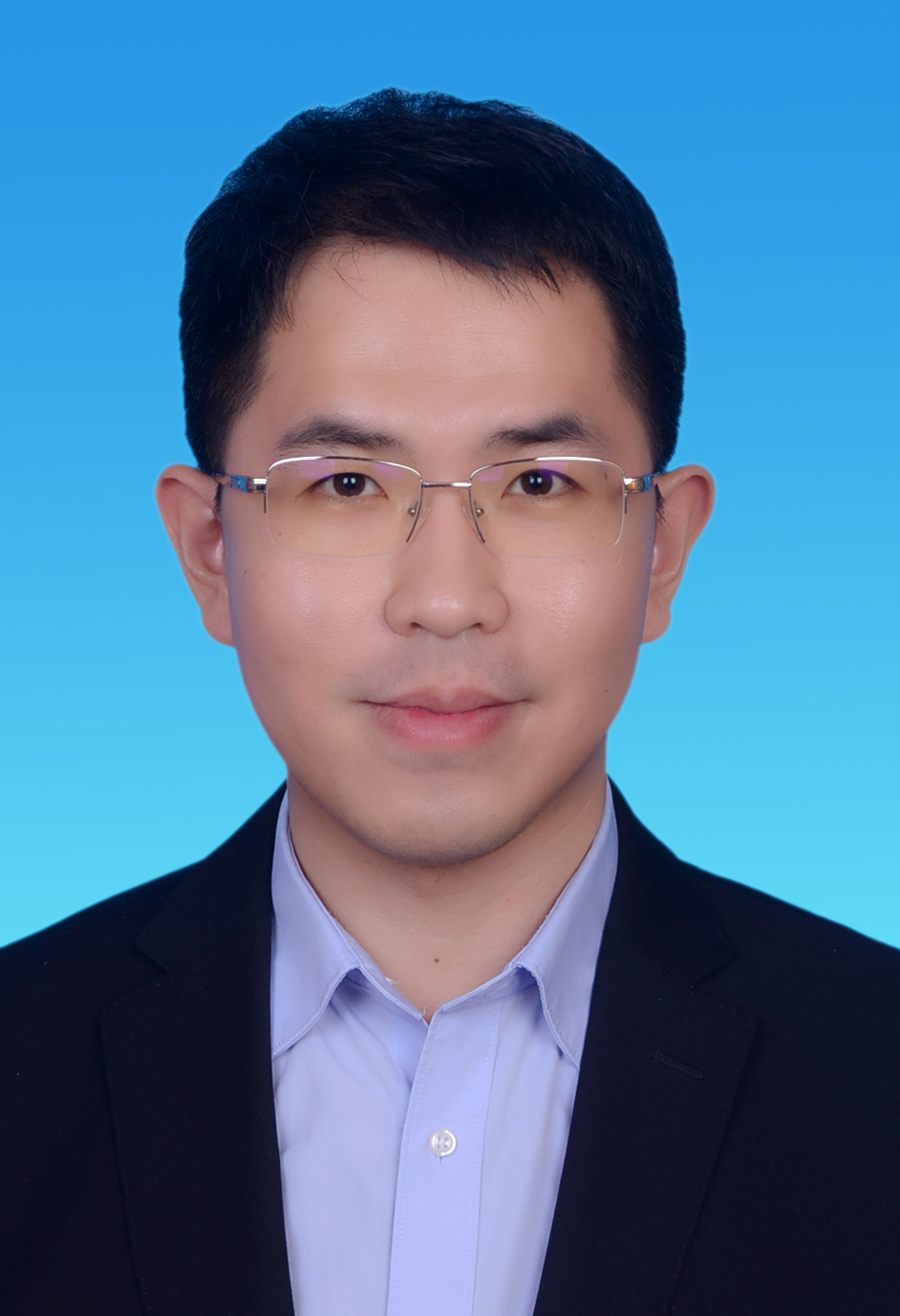}}]{Xin~Yang}
is a Professor in the Department of Computer Science at Dalian University of Technology, China. Yang received his B.S. degree in Computer Science from Jilin University in 2007. From 2007 to June 2012, he was a joint Ph.D. student at Zhejiang University and UC Davis for Graphics and received his Ph.D. degree in July 2012. His research interests include computer graphics and robotic vision.
\end{IEEEbiography}

\end{document}